\documentclass[letterpaper]{article} 
\usepackage{aaai2027}
\usepackage[hyphens]{url} 
\usepackage{graphicx}
\usepackage{natbib}  
\usepackage{caption}
\usepackage{placeins}
\usepackage{algorithm}
\usepackage{algorithmic}
\usepackage{xcolor}

\usepackage[table]{xcolor}
\definecolor{MetricBlue}{RGB}{236,245,249}

\usepackage{multirow}
\usepackage{array}

\usepackage{tabularx}
\usepackage{array}
\usepackage[table]{xcolor}

\usepackage{colortbl}
\definecolor{GroupGray}{gray}{0.94}

\newcommand{\brandicon}[1]{%
  \raisebox{-0.13em}{%
    \includegraphics[height=0.90em]{Figures/logos/#1}}%
  \hspace{0.35em}%
}
\usepackage{newfloat}
\usepackage{listings}
\DeclareCaptionStyle{ruled}{labelfont=normalfont,labelsep=colon,strut=off} 
\floatstyle{ruled}
\newfloat{listing}{tb}{lst}{}
\floatname{listing}{Listing}
\usepackage{booktabs}
\nocopyright

\renewcommand{\thesection}{\Roman{section}}

\makeatletter
\renewcommand{\p@subsection}{\thesection.}
\makeatother
\title{AesCanvas: A Large-Scale Dataset and Benchmark for Aesthetic Critique and Contextual Suitability}
\author{
    Xuanwei Hu\textsuperscript{\rm 1}\equalcontrib,
    Haoyu Dong\textsuperscript{\rm 1}\equalcontrib,
    Kejun Wu\textsuperscript{\rm 1}\corresponding,
    Tianyi Liu\textsuperscript{\rm 2},
    Jianjun Gao\textsuperscript{\rm 2}
}

\affiliations{
    \textsuperscript{\rm 1}School of Electronic Information and Communications,
    Huazhong University of Science and Technology, Wuhan, China\\
    \textsuperscript{\rm 2}School of Electrical and Electronic Engineering,
    Nanyang Technological University, Singapore\\
    kjwu@hust.edu.cn
}

\begin{document}

\maketitle

\begin{abstract}

Recent advances in Multimodal Large Language Models (MLLMs) have extended
Image Aesthetic Assessment (IAA) beyond scalar scores toward interpretable
critique and guidance. Yet existing benchmarks mainly assess intrinsic visual
quality or fixed domain criteria, leaving open whether an appealing image is
appropriate for a specific purpose, audience, cultural setting, or domain
convention. We introduce \textsc{AesCanvas}, a unified suite with two
complementary components: \textsc{CritiqueCanvas} with
519,136 instruction--response pairs from 54,300 images supports long-form,
multi-dimensional critique across photography, painting, and virtual imagery,
whereas \textsc{ContextCanvas} with 301 expert-reviewed use scenarios evaluates contextual aesthetic suitability in
realistic use scenarios. Under a unified protocol, we evaluate closed-source
frontier, open-weight general, and aesthetic-specific MLLMs. Results reveal a
clear separation between critique generation and context-sensitive judgment:
reference-based lexical and semantic metrics only partially capture critique
quality, while aesthetic specialists remain competitive on selected critique
metrics yet substantially lag strong general-purpose MLLMs on
\textsc{ContextCanvas}. Further analyses show that aesthetic specialization
does not reliably transfer to contextual suitability and that model decisions
may fail to track or ground themselves in decisive contextual visual cues.
These findings establish culturally situated, evidence-grounded suitability
as a distinct objective for aesthetic modeling.

\end{abstract}

\section{Introduction}

Image Aesthetic Assessment (IAA) has progressed from scalar prediction and
preference modeling~\cite{murray2012ava,talebi2018nima,yi2023artistic} toward
language-based aesthetic perception, critique, diagnosis, and
guidance~\cite{huang2024aesbench,huang2024aesexpertmultimodalityfoundationmodel,zhou2024uniaa,zhang2025Teachingb,qi2025photographer,
cao2025artimusefinegrainedimageaesthetics}. In practical use, however, visual appeal
alone is insufficient: an image must also suit a particular purpose, audience,
and convention. Cultural knowledge is therefore not auxiliary to aesthetic
judgment, since symbols, styles, dress, and genre conventions can directly
change whether an image is an appropriate visual choice. Related multimodal
studies likewise show that models remain fragile when judgments require
criterion-sensitive reasoning, visual grounding, or culturally situated
evidence~\cite{xiong2026multicrit,li2026groundingme,wu2026groundedcot,
satar2025seeingculture,singh2026curve}.

\begin{figure}[t]
    \centering
    \includegraphics[width=\columnwidth]{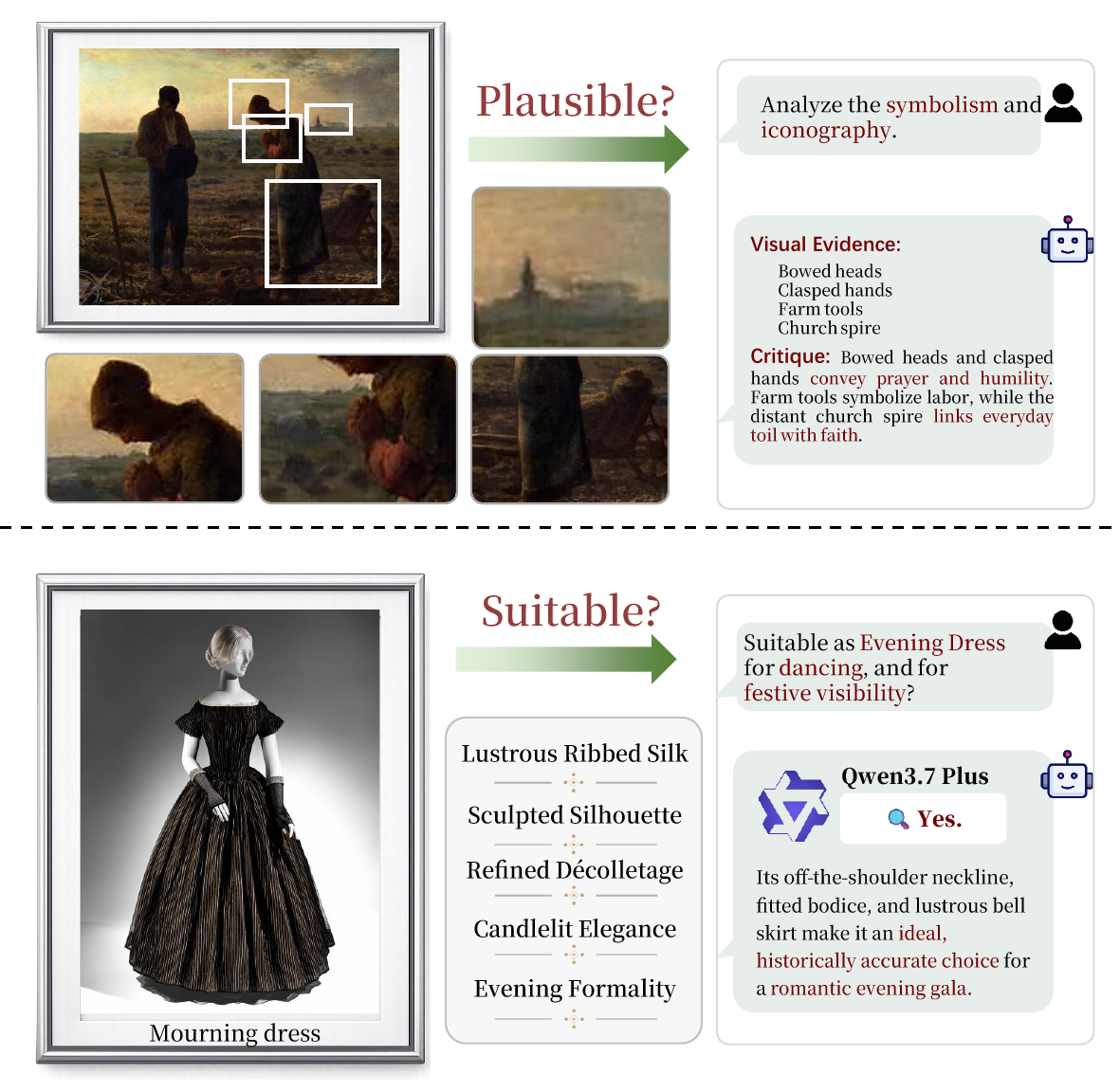}
    \caption{\textbf{Fluent aesthetic critique does not guarantee contextually valid judgment.} The top example shows an evidence-grounded critique, whereas the bottom shows a suitability error in which visually appealing attributes are mistaken for appropriateness in a mismatched use context. This contrast motivates evaluating critique grounding and contextual suitability as complementary capabilities.}
    \label{fig:contextcanvas-overview}
\end{figure}

\begin{figure*}[t]
    \centering
    \includegraphics[width=\textwidth]{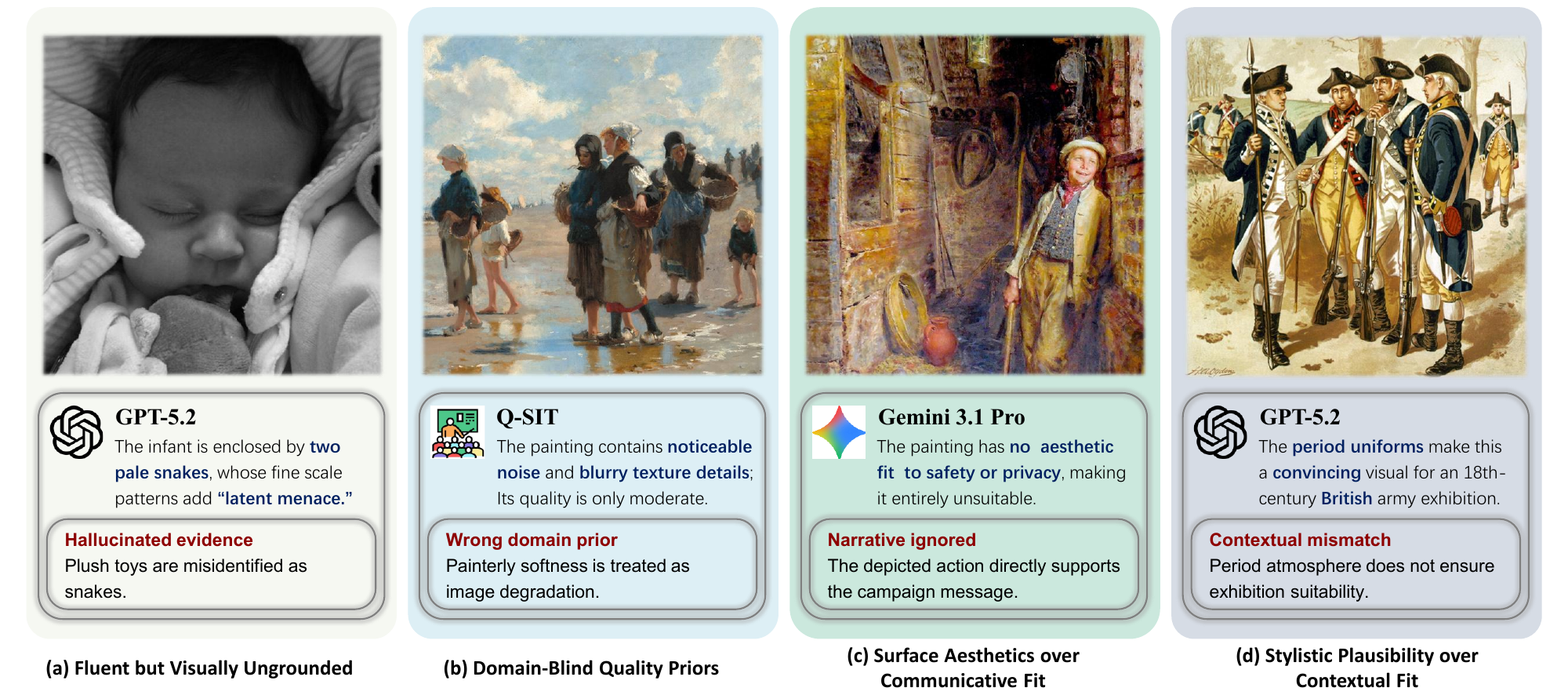}
    \caption{\textbf{Four recurring failure modes in practical aesthetic evaluation.}
    Models may generate critiques that are visually ungrounded or rely on
    inappropriate domain priors, and may judge real-world suitability from
    surface aesthetics or stylistic plausibility rather than communicative
    and contextual fit.}
    \label{fig:failure-modes}
\end{figure*}

Figure~\ref{fig:contextcanvas-overview} makes this distinction concrete: a
model may produce a plausible aesthetic critique yet approve a mourning dress
as festive evening wear by emphasizing its silhouette and material while
ignoring its social function. Figure~\ref{fig:failure-modes} further reveals
two failure levels: critiques may rely on fabricated evidence or inappropriate
domain priors, while suitability judgments may privilege surface appeal or
stylistic plausibility over cultural fit. We call the latter tendency
\emph{aesthetic context bias}: following generic aesthetic priors while

\FloatBarrier

\noindent neglecting contextual evidence that should alter the decision. This task is
not generic cultural question answering; contextual knowledge matters because
it determines whether the image works as a visual choice.

To study both capabilities, we introduce \textsc{AesCanvas}, a unified suite
with two complementary components. \textsc{CritiqueCanvas} provides
large-scale, multi-domain supervision for long-form critique across
photography, painting, and virtual imagery, while \textsc{ContextCanvas}
contains expert-reviewed use scenarios that distinguish visual appeal from
contextual suitability. Across closed-source, open-weight, and
aesthetic-specific MLLMs, we observe a clear capability gap: aesthetic
specialists remain competitive on selected critique metrics, yet all score
below 30\% on \textsc{ContextCanvas}, compared with 91.69\% for the strongest
evaluated general-purpose model. Further analyses show that reference-based
metrics only partially capture critique quality, aesthetic specialization does
not reliably transfer to contextual judgment, and correct decisions may still
lack grounding in decisive visual evidence.

\noindent\textbf{Contributions.}
\textbf{(1)} We introduce \textsc{CritiqueCanvas}, comprising 519,136 aesthetic instruction--response pairs built from 54,300 multi-domain images for instruction-tuning MLLMs on fine-grained image aesthetic assessment;
\textbf{(2)} we construct \textsc{ContextCanvas}, a 301-case expert-curated benchmark for contextual aesthetic suitability judgment;
\textbf{(3)} we evaluate closed-source, open-weight, and aesthetic-specific MLLMs, revealing a clear gap between producing fluent aesthetic critiques and correctly judging suitability under cultural, functional, and domain-specific constraints; and
\textbf{(4)} we release the data, prompts, rationales, provenance records, and evaluation code.

\section{Related Work}

\begin{figure*}[t]
    \centering
    \includegraphics[width=\textwidth]{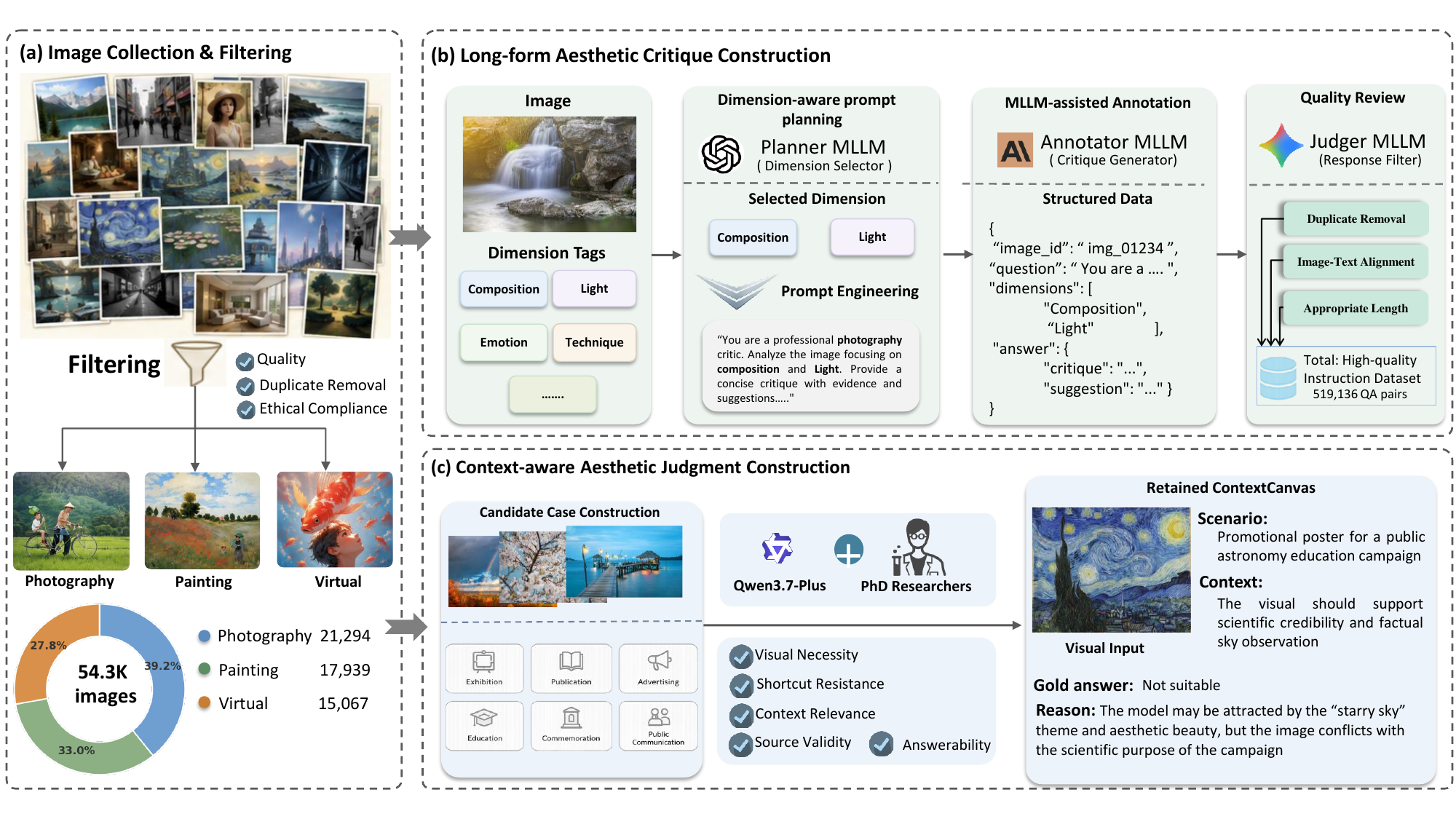}
    \caption{\textbf{Construction of \textsc{AesCanvas}.} 
    (a) A shared multi-domain image pool is collected and filtered. 
    (b) \textsc{CritiqueCanvas} is generated through dimension-aware prompt planning, MLLM-assisted annotation, and quality filtering. 
    (c) \textsc{ContextCanvas} is constructed from realistic aesthetic-use scenarios followed by expert review.}
    \label{fig:dataset_construction}
\end{figure*}

\subsection{Image Aesthetic Assessment}

Image Aesthetic Assessment (IAA) predicts human judgments of visual aesthetic
quality. Early work formulated IAA as classification, ranking, or
score-distribution prediction from crowd preferences, while incorporating
photographic attributes, composition-preserving representations, and
individual taste variation~\cite{murray2012ava,kong2016photo,
mai2016composition,yang2022personalized}. Recent studies extend IAA to artistic
images, fine-grained comparison within image series, and open-world criteria
that vary across themes~\cite{yi2023artistic,yang2026fine,liao2026open}.
Nevertheless, these methods primarily assess aesthetic strength under
intrinsic or task-fixed criteria, rather than whether an image is appropriate
for a particular purpose, audience, cultural setting, or domain convention.
Our work addresses this complementary problem of contextual aesthetic
suitability.

\subsection{Aesthetic Multimodal Models}

Aesthetic multimodal models extend conventional IAA from scalar prediction to
language-based recognition, description, interpretation, scoring, and
critique. Early studies benchmarked the aesthetic perception of general-purpose
MLLMs and explored aesthetic instruction tuning~\cite{huang2024aesbench,
huang2024aesexpertmultimodalityfoundationmodel,zhou2024uniaa}. Subsequent work connected
textual supervision with visual scoring through language-defined rating levels
or joint scoring-and-interpretation objectives~\cite{wu2024qalign,
zhang2025Teachingb}, while more recent models support professional critique,
multi-attribute analysis, and unified perceptual assessment~\cite{
qi2025photographer,cao2025artimusefinegrainedimageaesthetics,cao2025uniperceptunifiedperceptuallevelimage}. Existing evaluation,
however, remains centered on aesthetic perception, scoring, reference-aligned
critique, or domain-specific analysis. \textsc{AesCanvas} instead examines
whether aesthetic articulation transfers to contextual decisions in which an
attractive or stylistically plausible image may still conflict with its
intended use. Beyond conventional visual inputs, recent multimodal foundation models have also explored semantic understanding and language generation over specialized or corrupted visual representations~\cite{wu2026corrupted,liang2026cibic}.

\subsection{Contextual Multimodal Reasoning}

Contextual multimodal reasoning requires models to combine visual evidence
with task conditions, external knowledge, cultural conventions, and
criterion-specific constraints. Recent benchmarks study real-world visual
reasoning, context-dependent functional inference, and flexible judgment under
varying criteria~\cite{yin2026worldbench,zhang2026sfi,xiong2026multicrit,gao2024contextual}.
Grounding-focused work further evaluates whether intermediate reasoning and
final answers are supported by localized image evidence~\cite{
li2026groundingme,wu2026groundedcot,cai2024temporal}, while other studies expose failures under
visual illusions, counterfactual or adversarial evidence, and conflicts
between knowledge and images~\cite{hou2026viabench,zhou2026reactbench,
moratelli2026deflection}. Culturally situated benchmarks likewise reveal
fragility on region-specific artifacts and multilingual cultural
evidence~\cite{satar2025seeingculture,singh2026curve}. However, these works
primarily assess factual correctness, grounding, functional inference, or
adherence to specified criteria, rather than whether contextual evidence
should outweigh surface aesthetic appeal in a concrete visual-use decision.
\textsc{ContextCanvas} targets this intersection through use-oriented
aesthetic suitability judgments grounded in both visual and contextual
evidence.

\section{Dataset}

\subsection{Dataset Overview}

We introduce \textsc{AesCanvas}, a unified suite built from 54,300 images to study two complementary aspects of aesthetic intelligence: articulating visual qualities and judging their suitability in context. \textsc{CritiqueCanvas} contains 519,136 instruction--response pairs across photography, painting, and virtual imagery for long-form, multi-dimensional critique generation. \textsc{ContextCanvas} comprises 301 expert-designed and reviewed cases that evaluate contextual aesthetic suitability---whether an image is an appropriate visual choice for a specified purpose, audience, cultural setting, or domain convention, beyond its intrinsic visual appeal. Together, the two components test whether models that can identify and articulate aesthetic evidence can also translate it into contextually valid decisions.

\subsection{Dataset Curation}

As illustrated in Fig.~\ref{fig:dataset_construction}, our pipeline branches from a shared multi-domain image pool into \textsc{CritiqueCanvas} for structured aesthetic critique and \textsc{ContextCanvas} for closed-form contextual suitability judgment.

\paragraph{Image Collection and Filtering.}

We collect candidate images from three visual domains: \emph{photography} (e.g., natural scenes, architecture, portraiture, and commercial imagery), \emph{painting} (primarily Western classical and related works), and \emph{virtual imagery} (e.g., game scenes, animation-style images, and AI-generated content). Images are drawn from publicly accessible web sources and online museum and cultural-heritage collections, with available source URLs and provenance metadata retained. After removing low-quality,
semantically uninformative, non-compliant, and duplicate samples, the pool contains 21,294 photographs, 17,939 paintings, and 15,067 virtual images, for a total of 54,300.

\paragraph{\textsc{CritiqueCanvas} Annotation and Construction.}
As shown in Fig.~\ref{fig:dataset_construction}(b), we use a prompt-based pipeline that combines shared aesthetic dimensions with domain-specific criteria. The shared dimensions include \emph{Content and Narrative}, \emph{Composition}, \emph{Color}, \emph{Lighting}, \emph{Lines and Brushstrokes}, \emph{Style}, \emph{Emotion}, \emph{Technique}, \emph{Symbolism}, and \emph{Visual Appeal}. Annotating
models select only the dimensions most relevant to each image rather than addressing all dimensions uniformly.

Prompts are further conditioned on the visual domain. Photography emphasizes exposure, perspective, depth of field, focus, motion, and compositional control; painting emphasizes line, brushwork, genre, style, symbolism, and emotional expression; and virtual imagery emphasizes character design, scene
construction, rendering consistency, narrative setting, and stylization. Multiple MLLMs generate structured annotations, which are converted into multi-turn instruction--response conversations covering description, evaluation, interpretation, technical analysis, and constructive improvement.

An independent Claude Opus 5 verifier screens generated pairs for instruction
consistency, visual grounding, and response quality. A stratified audit of
1,000 retained pairs yields 95.3\% acceptance and 97\% raw agreement; full
criteria and statistics appear in the supplement.

\paragraph{\textsc{ContextCanvas} Construction and Review.}

As shown in Fig.~\ref{fig:dataset_construction}(c), \textsc{ContextCanvas} is constructed from a selected subset of the shared image pool. Four PhD-level
researchers with interdisciplinary backgrounds developed and reviewed 1,060 candidate cases. Each places an image in a plausible use scenario---such as
exhibition, publication, advertising, education, commemoration, or public communication---and asks whether it is an appropriate visual choice for the
stated purpose, audience, and context. Some cases include a curator- or designer-proposed assessment that the model must independently evaluate.

A case is retained only when it satisfies seven prespecified criteria: (1) \emph{aesthetic centrality}, requiring a judgment about the image as a visual choice; (2) \emph{context dependence}, such that intrinsic quality alone is insufficient; (3) \emph{visual necessity}, such that the task cannot
be reduced to text-only factual or cultural knowledge; (4) \emph{decisive evidence}, such that relevant cultural, historical, or narrative evidence materially affects the answer; (5) \emph{scenario naturalness}, requiring a plausible design or communication use; (6) \emph{answerability}, requiring a
unique closed-form answer from perceptible evidence; and (7) \emph{anti-shortcut validity}, preventing wording, option length, or label priors from revealing the answer.

All candidates were reviewed against these criteria, with borderline cases resolved through discussion. Qwen3.7-Plus was used only as a development-time
probe for trivial, ambiguous, or shortcut-prone cases; model failure alone never determined retention. All inclusion decisions, gold labels, rationales,
and source records were finalized solely by human reviewers, and its reported score is therefore marked as a development-model diagnostic.

The final benchmark contains 301 cases (28.40\%): 300 single-image and one
paired-image case, covering 302 images, 291 binary questions, and 10 three-way
questions. Each includes the visual input, use scenario, context, answer options, gold label, source-grounded rationale, and provenance. The
cases span painting, sculpture, illustration and comics, animation and digital
imagery, photography, and film.

\begin{table}[t]
\centering

\footnotesize
\setlength{\tabcolsep}{3.2pt}
\renewcommand{\arraystretch}{1.02}

\begin{tabular}{
@{}
>{\raggedright\arraybackslash}p{0.13\columnwidth}
>{\raggedright\arraybackslash}p{0.31\columnwidth}
>{\raggedright\arraybackslash}p{0.48\columnwidth}
@{}
}
\toprule
\textbf{Task} & \textbf{Metric Family} & \textbf{Metrics} \\
\midrule

\multirow[t]{4}{*}{\textsc{Critique}}
& Lexical overlap
& BLEU, ROUGE-L, METEOR \\
\cmidrule(lr){2-3}

& Semantic similarity
& BERT-F1, SBERT-Cos \\
\cmidrule(lr){2-3}

& Image--text alignment
& CLIPScore \\
\cmidrule(lr){2-3}

& Descriptive statistics
& Avg. Length, Avg. Dim. Hits, Top-3 Dimensions \\

\midrule

\multirow[t]{2}{*}{\textsc{Context}}
& Prediction quality
& Accuracy, Macro-F1 \\
\cmidrule(lr){2-3}

& Prediction tendency
& Yes Rate \\

\bottomrule
\end{tabular}
\caption{Metrics used for the two evaluation tasks.}
\label{tab:evaluation_metrics}
\end{table}

\subsection{Evaluation Protocol}
\label{sec:evaluation_protocol}

We evaluate critique generation on \textsc{CritiqueCanvas} and contextual
suitability on \textsc{ContextCanvas}. Models receive identical inputs and
instructions without web search, retrieval, or auxiliary recognition tools;
Table~\ref{tab:evaluation_metrics} summarizes the metrics.

\begin{table*}[!htbp]
\centering
{\small

\renewcommand{\arraystretch}{1.05}
\begin{tabularx}{0.74\textwidth}{
@{}
>{\raggedright\arraybackslash}X
>{\centering\arraybackslash}p{0.115\textwidth}
>{\centering\arraybackslash}p{0.115\textwidth}
>{\centering\arraybackslash}p{0.105\textwidth}
@{}
}
\toprule
\textbf{Model} & \textbf{Acc.} & \textbf{Macro-F1} & \textbf{Yes (\%)} \\
\midrule

\rowcolor{GroupGray}
\multicolumn{4}{c}{\textit{Closed-source MLLMs}} \\
\brandicon{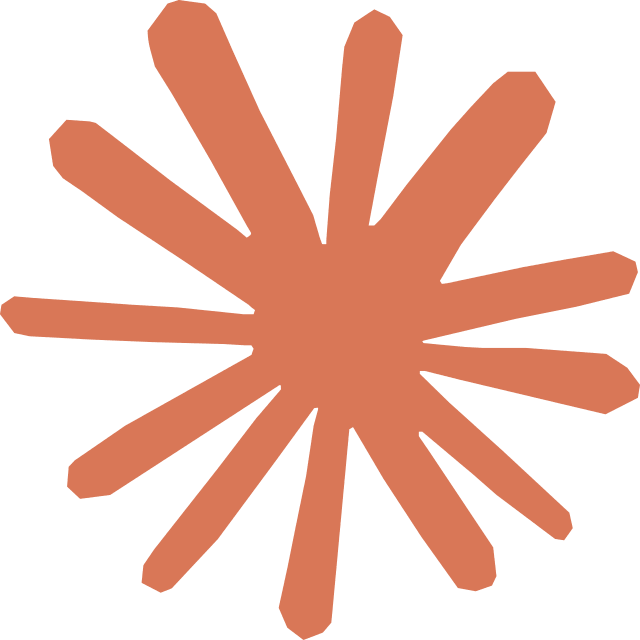}Claude Opus 5
  & \textbf{91.69} & \textbf{90.45} & 30.24 \\
\brandicon{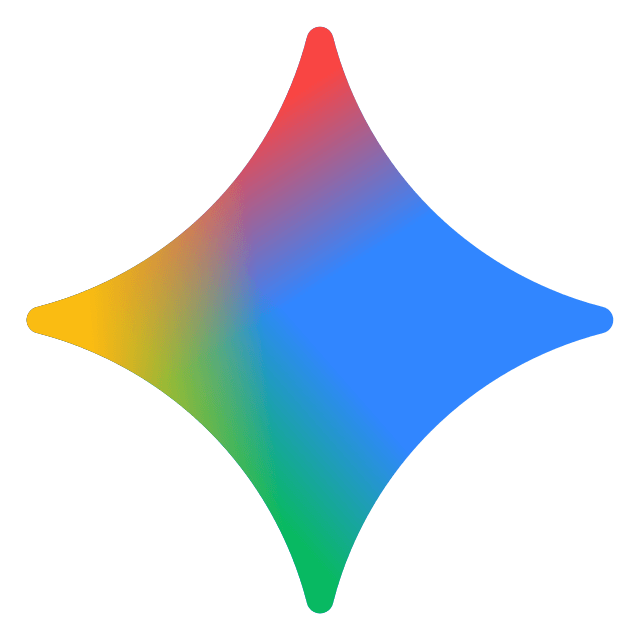}Gemini 3.1 Pro
  & 85.71 & 83.06 & 26.12 \\
\brandicon{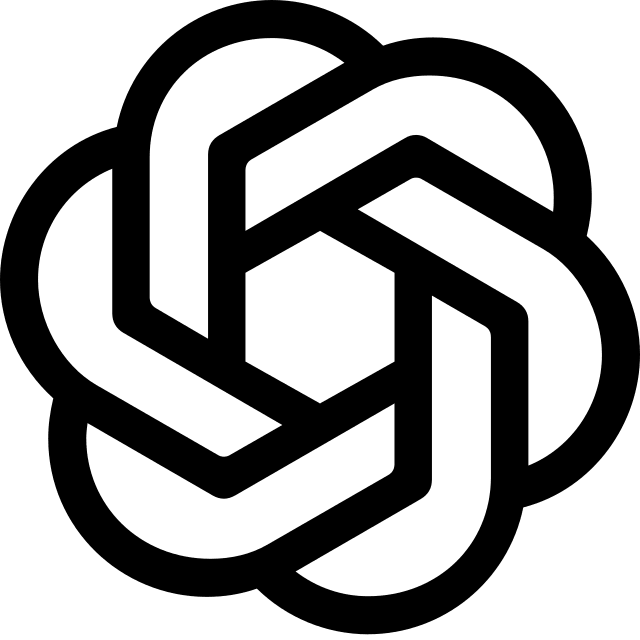}GPT-5.5
  & 83.06 & 80.13 & 27.49 \\
\brandicon{claude-color.png}Claude Opus 4.6
  & 72.76 & 67.96 & 27.15 \\
\brandicon{openai.png}GPT-5.2
  & 71.43 & 64.51 & 20.96 \\
\brandicon{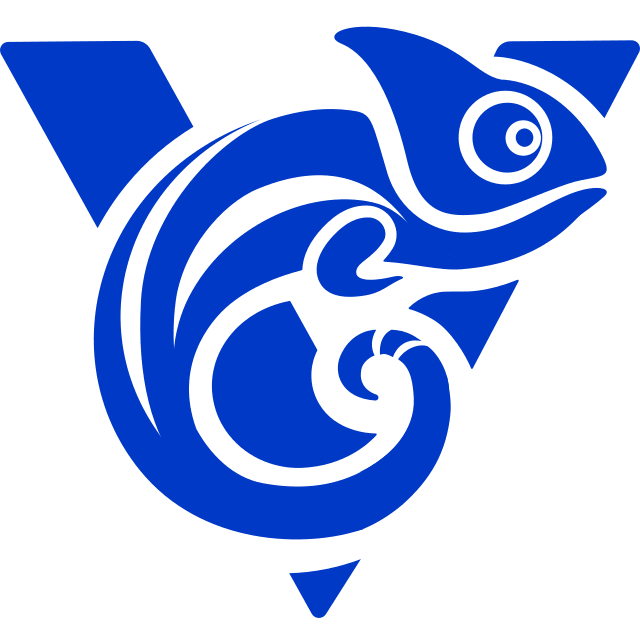}GLM-5V-Turbo
  & 57.81 & 55.22 & 43.64 \\
\brandicon{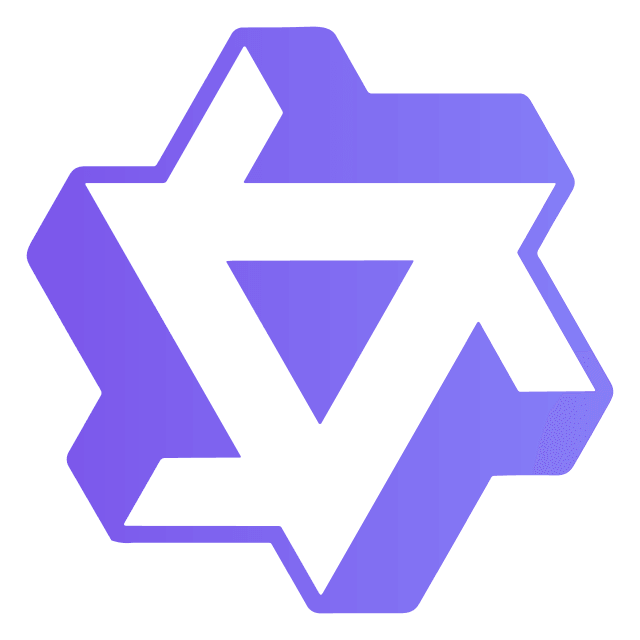}Qwen3.7-Plus$^{\dagger}$
  & 51.83 & 48.39 & 41.58 \\
\brandicon{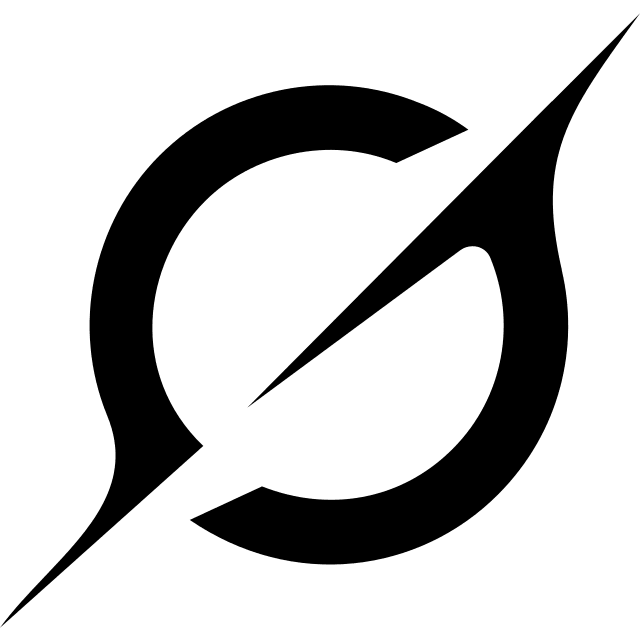}Grok 4.20
  & 44.19 & 39.09 & 37.80 \\
\midrule

\rowcolor{GroupGray}
\multicolumn{4}{c}{\textit{Open-weight General MLLMs}} \\
\brandicon{qwen-color.png}Qwen3-VL-Instruct ~\cite{bai2025Qwen3VL}         & 42.19 & 38.74 & 44.67 \\
\brandicon{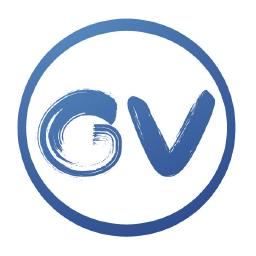}InternVL3 ~\cite{zhu2025InternVL3}                 & 23.26 & 21.65 & 62.20 \\
\brandicon{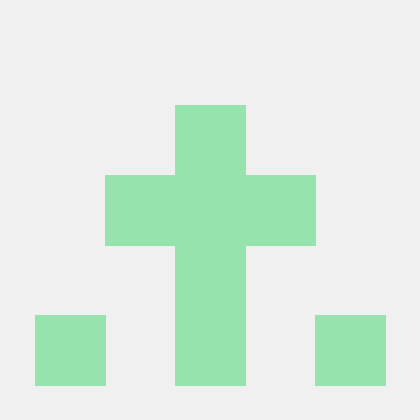}mPLUG-Owl2 ~\cite{ye2024mplug}                    & 20.27 & 17.43 & 50.17 \\
\brandicon{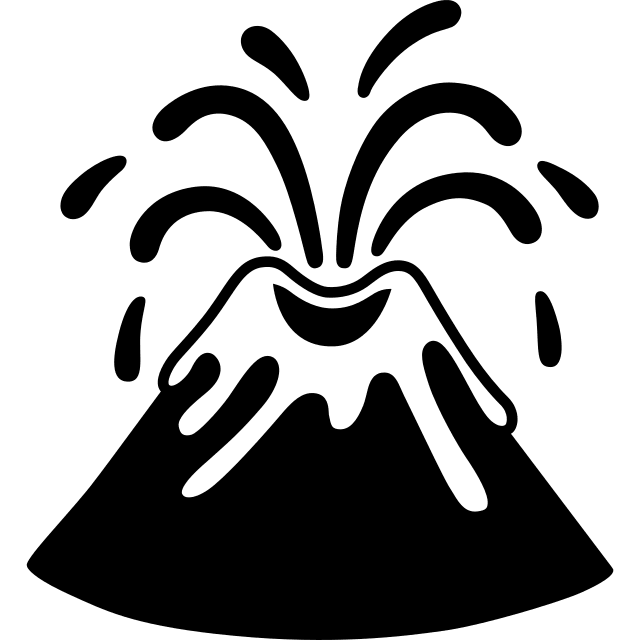}LLaVA-OneVision ~\cite{an2025LLaVAOneVision15}                        & 19.93 & 18.20 & 71.13 \\
\brandicon{llava.png}LLaVA-v1.5 ~\cite{liu2024improved}        & 12.62 & 10.29 & 60.48 \\
\midrule

\rowcolor{GroupGray}
\multicolumn{4}{c}{\textit{Aesthetic-specific Models}} \\
ArtQuant-APDD ~\cite{liuBridgingCognitiveGap2026} & 29.57 & 29.50 & 59.11 \\
ArtiMuse ~\cite{cao2025artimusefinegrainedimageaesthetics}     & 19.60 & 17.87 & 62.54 \\
Q-SiT ~\cite{zhang2025Teachingb}        & 19.60 & 16.65 & 76.98 \\
AesExpert ~\cite{huang2024aesexpertmultimodalityfoundationmodel}    & 18.27 & 16.30 & 66.67 \\
\bottomrule
\end{tabularx}}
\caption{Contextual aesthetic suitability results. Acc. is exact-match
accuracy on all 301 cases. Macro-F1 and Yes (\%) are computed on the 291
binary cases; the gold Yes rate is 38.14\%. Best Acc. and Macro-F1 are
bolded. $\dagger$ denotes the development model used for difficulty screening.}
\label{tab:diagnostic_results}
\end{table*}

\paragraph{Long-form Aesthetic Critique Generation.}
We evaluate models on 6,000 open-ended samples balanced across photography,
painting, and virtual imagery, requiring aesthetic analysis or image-grounded
improvement suggestions. Image-level splits keep related conversations together.

Because valid critiques may differ substantially in wording, we report
complementary metric families. BLEU ~\cite{papineni-etal-2002-bleu}, ROUGE-L ~\cite{lin-2004-rouge}, and METEOR ~\cite{banerjee-lavie-2005-meteor} measure lexical
overlap; BERT-F1 ~\cite{zhang2020bertscoreevaluatingtextgeneration} and SBERT-Cos ~\cite{reimers-gurevych-2019-sentence} measure token- and sentence-level semantic
similarity; and CLIPScore ~\cite{hessel-etal-2021-clipscore} measures image--text relevance. These metrics are
interpreted jointly rather than as complete measures of aesthetic reasoning.

To further characterize response behavior, we report average response length,
average aesthetic-dimension coverage, and the three most frequently addressed
dimensions in the supplementary material. Specifically, let $\mathcal{D}$
denote the predefined set of aesthetic dimensions, and let
$m(\hat{r}_i,d)$ indicate whether response $\hat{r}_i$ explicitly addresses
dimension $d$. The average number of dimensions addressed per response is

\begin{equation}
\mathrm{DimHits}
=
\frac{1}{N}
\sum_{i=1}^{N}
\sum_{d \in \mathcal{D}}
m(\hat{r}_i,d).
\end{equation}

This statistic measures the breadth of explicit aesthetic coverage rather than
its correctness: a higher value does not necessarily indicate that the
identified dimensions are relevant, accurate, or well grounded.

\paragraph{Contextual Aesthetic Suitability Judgment.}
For each \textsc{ContextCanvas} case, the model receives an image, a realistic
use scenario, and fixed answer options, and returns a single option label.
Explanations are retained for qualitative analysis but do not affect the
primary score. A prediction is correct only if a unique option is parsed and
matches the gold answer; missing, conflicting, or unparsable responses are
counted as incorrect.

We report exact-match accuracy over all 301 cases. Macro-F1 and predicted
Yes Rate are computed on the 291 binary cases after mapping the options to
canonical \emph{Yes}/\emph{No} labels; the remaining 10 three-way cases are
included only in accuracy. The gold Yes Rate is 38.14\%, against which the
predicted rate indicates a model's tendency toward over-acceptance or
over-rejection.

\section{Experiments}

\subsection{Experimental Setup}
\label{sec:experimental_setup}

\paragraph{Evaluated Models.}
We compare closed-source frontier MLLMs, open-weight general MLLMs, and
aesthetic-specific models, separating general visual-language capability from
specialization for aesthetic assessment or critique. Exact model variants are
listed in Tables~\ref{tab:diagnostic_results} and~\ref{tab:benchmark_results}. Qwen3-VL$_{\mathrm{FT}}$ denotes a variant
adapted on \textsc{CritiqueCanvas} and evaluated only on critique generation.

\paragraph{Implementation Details.}
All models follow the protocol in Section~\ref{sec:evaluation_protocol}.
Locally deployed models receive images resized or padded to
$448\times448$ pixels. We use deterministic decoding whenever supported and
a maximum of 256 tokens for critique generation; model-specific settings are
reported in the supplementary material.

\begin{table*}[!htbp]
\centering
{\small
\setlength{\tabcolsep}{5.4pt}
\renewcommand{\arraystretch}{1.06}

\begin{tabular*}{\textwidth}{@{\extracolsep{\fill}}lcccccc@{}}
\toprule
\textbf{Model} & \textbf{BLEU} & \textbf{ROUGE-L} & \textbf{METEOR} & \textbf{BERT-F1} & \textbf{SBERT-Cos} & \textbf{CLIPScore} \\
\midrule
\rowcolor{GroupGray}
\multicolumn{7}{c}{\textit{Closed-source MLLMs}} \\
\brandicon{gemini-color.png}Gemini 3.1 Pro & 7.15 & 0.250 & 0.281 & 0.699 & \textbf{0.848} & 0.303 \\
\brandicon{openai.png}GPT-5.2 & 2.84 & 0.210 & 0.209 & 0.669 & 0.818 & 0.301 \\
\midrule
\rowcolor{GroupGray}
\multicolumn{7}{c}{\textit{Open-weight General MLLMs}} \\
\brandicon{glmv-color.png}GLM-4.6-Flash ~\cite{team2026glm45v}          & 5.50 & 0.241 & 0.273 & 0.676 & 0.844 & 0.317 \\
\brandicon{llava.png}LLaVA-OneVision ~\cite{an2025LLaVAOneVision15}     & 10.32 & 0.285 & 0.275 & \textbf{0.724} & 0.839 & \textbf{0.319} \\
\brandicon{internvl.png}InternVL ~\cite{wang2025InternVL35}             & 9.15 & 0.272 & 0.280 & 0.716 & 0.829 & 0.300 \\
\brandicon{qwen-color.png}Qwen3-VL ~\cite{bai2025Qwen3VL}               & 5.12 & 0.232 & 0.274 & 0.684 & 0.827 & 0.313 \\
\brandicon{qwen-color.png}Qwen3-VL$_{\mathrm{FT}}$                   & \textbf{12.23} & \textbf{0.297} & \textbf{0.296} & 0.722 & 0.787 & 0.285 \\
\midrule
\rowcolor{GroupGray}
\multicolumn{7}{c}{\textit{Aesthetic-specific Models}} \\
ArtQuant-APDD ~\cite{liuBridgingCognitiveGap2026} & 10.68 & 0.280 & 0.277 & 0.716 & 0.818 & 0.297 \\
ArtiMuse ~\cite{cao2025artimusefinegrainedimageaesthetics} & 7.98 & 0.271 & 0.239 & 0.718 & 0.826 & 0.312 \\
UniPercept ~\cite{cao2025uniperceptunifiedperceptuallevelimage} & 5.12 & 0.235 & 0.200 & 0.693 & 0.823 & 0.318 \\
AesExpert ~\cite{huang2024aesexpertmultimodalityfoundationmodel} & 2.08 & 0.198 & 0.128 & 0.651 & 0.609 & 0.262 \\
Q-SiT ~\cite{zhang2025Teachingb} & 1.31 & 0.188 & 0.108 & 0.635 & 0.508 & 0.261 \\
\bottomrule
\end{tabular*}}

\caption{
Quantitative evaluation of long-form aesthetic critique generation on
CritiqueCanvas. The reported metrics assess complementary aspects of
generated critiques, including reference alignment through lexical and
semantic similarity, and image--text relevance.
}
\label{tab:benchmark_results}
\end{table*}

\subsection{Main Results}

\subsubsection{Contextual Aesthetic Suitability Judgment}

Table~\ref{tab:diagnostic_results} reports results on \textsc{ContextCanvas}. The benchmark clearly separates model capability levels. Accuracy among closed-source MLLMs ranges from $44.19\%$ to
$91.69\%$, while the strongest evaluated open-weight model, Qwen3-VL-Instruct, reaches only $42.19\%$. Within the Claude and GPT families, newer variants substantially outperform their predecessors: Claude Opus 5 improves over Claude Opus 4.6 by $18.93$ accuracy points and $22.49$ Macro-F1 points, while GPT-5.5 improves over GPT-5.2 by $11.63$ and $15.62$ points, respectively. These within-family gains are consistent with improved contextual
aesthetic judgment in newer frontier models.

Most notably, all evaluated aesthetic-specific models achieve below $30\%$ accuracy. ArtQuant-APDD, the strongest model in this group, obtains $29.57\%$, compared with $42.19\%$ for Qwen3-VL-Instruct and
$91.69\%$ for Claude Opus 5. ArtiMuse, Q-SiT, and AesExpert achieve only $19.60\%$, $19.60\%$, and $18.27\%$, respectively. Thus, specialization for aesthetic scoring, perception, or critique does not reliably transfer to judging suitability under cultural, communicative, or domain-specific constraints.

The Yes-rate column further reveals a shared prediction tendency. Against the gold rate of $38.14\%$, aesthetic-specific models predict Yes in $59.11\%$--$76.98\%$ of the binary cases, indicating pronounced
over-acceptance despite contextual mismatch. This pattern is consistent with aesthetic context bias, although similar marginal rates may arise from different instance-level failures. We examine these failure modes through category-level and qualitative analyses in the following section. Although the binary subset has a $61.86\%$ always-\emph{No} majority baseline, weaker models often fall below it because they over-accept visually plausible but contextually unsuitable images. Meanwhile, frontier models achieve up to $91.69\%$ accuracy on the full benchmark, suggesting that \textsc{ContextCanvas} rewards context-sensitive judgment rather than majority-label guessing.

\subsubsection{Long-form Aesthetic Critique Generation}

Table~\ref{tab:benchmark_results} reports results on \textsc{CritiqueCanvas}. No model dominates across all metric families. Gemini 3.1 Pro achieves the highest SBERT-Cos ($0.848$), while LLaVA-OneVision leads BERT-F1 ($0.724$) and CLIPScore ($0.319$). The adapted Qwen3-VL$_{\mathrm{FT}}$ instead obtains the strongest
lexical scores: $12.23$ BLEU, $0.297$ ROUGE-L, and $0.296$ METEOR. Compared with the base Qwen3-VL, these correspond to gains of $7.11$, $0.065$, and $0.022$, respectively. However, its SBERT-Cos decreases
from $0.827$ to $0.787$, and CLIPScore from $0.313$ to $0.285$, showing that adaptation improves reference-style matching without uniform gains in semantic similarity or image grounding.

Aesthetic-specific models remain competitive on selected critique metrics. ArtiMuse reaches $0.718$ BERT-F1, close to the best score of $0.724$, while UniPercept obtains a CLIPScore of $0.318$, only $0.001$ below the overall best. ArtQuant also reaches $10.68$ BLEU and $0.280$ ROUGE-L, whereas AesExpert and Q-SiT perform weakly across most metrics. Overall critique performance is heterogeneous across lexical, semantic, and image-grounding criteria.

Together, the two tasks reveal a gap between reference-aligned critique
generation and contextually valid aesthetic judgment. Additional response-level
statistics are provided in the supplementary material.

\subsection{Analysis}

Beyond aggregate scores, we examine why aesthetic competence
fails to transfer to culturally situated judgment. In the compact
tables, GPT, Gem., Qwen+, Qwen-FT, AQ, and AM abbreviate
GPT-5.2, Gemini 3.1 Pro, Qwen3.7-Plus, the
\textsc{CritiqueCanvas}-adapted Qwen3-VL, ArtQuant-APDD, and
ArtiMuse, respectively; vertical rules follow the model groups in
Section~\ref{sec:experimental_setup}.

\begin{table}[!htbp]
\centering

{\footnotesize
\setlength{\tabcolsep}{3.2pt}
\renewcommand{\arraystretch}{1.05}
\begin{tabular*}{\columnwidth}{
@{\extracolsep{\fill}}l|cc|c|cc@{}
}
\toprule
\textbf{Rating}
& \textbf{GPT}
& \textbf{Gem.}
& \textbf{Qwen3-VL$_{\mathrm{FT}}$}
& \textbf{AM}
& \textbf{AQ} \\
\midrule
Human & 3.9 & \textbf{4.1} & 2.5 & 2.6 & 2.3 \\
MLLM  & \textbf{5.0} & 4.5 & 2.5 & 2.6 & 2.2 \\
\bottomrule
\end{tabular*}
}

\caption{Human and MLLM ratings of overall critique quality.}
\label{tab:overall_critique_rating}

{\footnotesize
\setlength{\tabcolsep}{3.2pt}
\renewcommand{\arraystretch}{1.05}
\begin{tabular*}{\columnwidth}{
@{\extracolsep{\fill}}l|ccc|cc@{}
}
\toprule
\textbf{Metric}
& \textbf{GPT}
& \textbf{Gem.}
& \textbf{Qwen+}
& \textbf{AQ}
& \textbf{AM} \\
\midrule
Correct $\uparrow$
& \textbf{13} & 10 & 11 & 3 & 6 \\
Reverse $\downarrow$
& 2 & 1 & 1 & 2 & \textbf{0} \\
NCU $\uparrow$
& \textbf{30.56} & 25.00 & 27.78 & 2.78 & 16.67 \\
\bottomrule
\end{tabular*}

\caption{Results of the 36-pair counterfactual audit.}
\label{tab:counterfactual}
}

\end{table}

\subsubsection{Validity of Critique Metrics}

We evaluate whether reference-based metrics reflect practical critique quality on 100 randomly sampled \textsc{CritiqueCanvas} cases. 2 human evaluators rate five representative models on visual grounding, aesthetic specificity, criterion appropriateness, and analytical usefulness, with Claude Opus 5 as the MLLM judge
providing a complementary assessment.

As shown in Table~\ref{tab:overall_critique_rating}, direct quality ratings do not consistently track the reference-similarity metrics in Table~\ref{tab:benchmark_results}. Since valid critiques may emphasize
different evidence and interpretations, lexical and semantic overlap capture only part of critique quality and should be complemented by direct judgments of grounding, specificity, and usefulness.

\subsubsection{Culture-Grounded Counterfactual Sensitivity}

We further examine whether contextual aesthetic decisions are grounded in
culture-bearing visual evidence. We construct 36 paired counterfactual
variants from ContextCanvas, keeping the evaluation scenario and overall
visual presentation fixed while changing the decisive cue such that the
intended judgment changes from \textit{No} to \textit{Yes}. These synthetic
variants neither revise the original annotations nor constitute an additional
accuracy split; they are used only as a matched diagnostic of directional
decision updating. We report the numbers of correct (\textit{No}$\rightarrow$
\textit{Yes}) and reverse (\textit{Yes}$\rightarrow$\textit{No}) updates, as
well as Net Correct Update computed as $100 \times (N_{\mathrm{correct}} - N_{\mathrm{reverse}})/36$.

As shown in Table~\ref{tab:counterfactual}, all three general-purpose MLLMs show positive net updating, with NCU values concentrated
between 25.00 and 30.56 despite their substantially different benchmark
accuracies. This suggests that absolute contextual competence and
responsiveness to decisive visual evidence are related but separable.
More importantly, both evaluated aesthetic specialists show weaker net
updating than every general-purpose model, with ArtQuant exhibiting almost
no measurable response. The results indicate that aesthetic specialization
alone does not establish the visual--cultural binding required for reliable
contextual aesthetic judgment.

\subsubsection{Transfer from Aesthetic Specialization}

To test whether specialization for conventional aesthetic tasks transfers to
contextual suitability judgment, we compare each aesthetic specialist with its
corresponding base model on the same \textsc{ContextCanvas} cases.

\begin{figure}[t]
\centering
\includegraphics[width=\columnwidth]{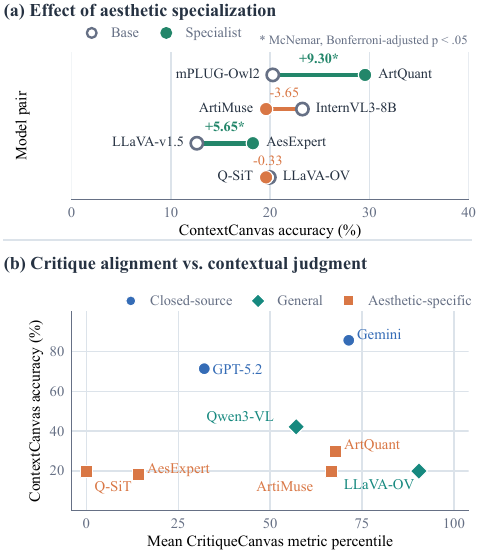}
\caption{\textbf{Cross-task analysis.} (a) Base--specialist transfer on
ContextCanvas; * denotes Bonferroni-corrected McNemar $p<.05$.
(b) ContextCanvas accuracy against the mean percentile over six
CritiqueCanvas metrics.}
\label{fig:cross_task}
\end{figure}

As shown in Figure~\ref{fig:cross_task}, specialization significantly
improves ArtQuant and AesExpert over their respective base models, but
yields no measurable gain for ArtiMuse or Q-SiT. Moreover, strong
CritiqueCanvas metric alignment does not consistently translate into
ContextCanvas accuracy.

\begin{table}[!htbp]
\centering

{\footnotesize
\setlength{\tabcolsep}{3.2pt}
\renewcommand{\arraystretch}{1.05}
\begin{tabular*}{\columnwidth}{
@{\extracolsep{\fill}}l|ccc|c@{}
}
\toprule
\textbf{Metric}
& \textbf{Gem.}
& \textbf{GPT}
& \textbf{Qwen+}
& \textbf{AM} \\
\midrule
Accuracy $\uparrow$
& \textbf{83} & 65 & 42 & 21 \\
EGA $\uparrow$
& \textbf{79} & 55 & 31 & 3 \\
Unsupported $\downarrow$
& \textbf{4.82} & 15.38 & 26.19 & 85.71 \\
\bottomrule
\end{tabular*}
}

\caption{Decisive-cue attribution on 100 balanced cases.}
\label{tab:ega_results}

\end{table}

\subsubsection{Decisive-Cue Grounding}

Binary accuracy does not show whether a model is correct for the right visual
reason. We therefore evaluate 100 balanced \textsc{ContextCanvas} cases using
human judgments of answer correctness, decisive-cue identification, and
contextual explanation. Evidence-Grounded Accuracy (EGA) requires all three.

As shown in Table~\ref{tab:ega_results}, general-purpose models preserve much
of their accuracy under the stricter grounding criterion, whereas ArtiMuse
drops from 21\% accuracy to 3\% EGA. Unsupported is the percentage of correct predictions that fail the
decisive-cue grounding criterion. This gap indicates that fluent aesthetic
rationales may still fail to identify the culture-bearing evidence that
determines contextual suitability.

\section{Conclusion}

We introduced \textsc{AesCanvas}, comprising \textsc{CritiqueCanvas} and
\textsc{ContextCanvas}, to evaluate aesthetic critique and contextual
suitability. Across closed-source, open-weight, and aesthetic-specific
models, strong critique performance does not reliably translate into
context-sensitive judgment: aesthetic specialists may remain competitive
on critique metrics yet achieve low accuracy and exhibit over-acceptance on
\textsc{ContextCanvas}. Paired and diagnostic analyses further reveal
inconsistent benefits from aesthetic tuning and limited use of decisive
contextual evidence. These findings support treating culturally situated
suitability as a distinct training and evaluation objective that requires
models to integrate visual evidence with cultural knowledge, intended use,
and domain-specific constraints.
\bibliography{ref}

@misc{caoArtiMuseFineGrainedImage2025a,
  title = {{{ArtiMuse}}: {{Fine-Grained Image Aesthetics Assessment}} with {{Joint Scoring}} and {{Expert-Level Understanding}}},
  author = {Cao, Shuo and Ma, Nan and Li, Jiayang and Li, Xiaohui and Shao, Lihao and Zhu, Kaiwen and Zhou, Yu and Pu, Yuandong and Wu, Jiarui and Wang, Jiaquan and Qu, Bo and Wang, Wenhai and Qiao, Yu and Yao, Dajuin and Liu, Yihao},
  year = 2025,
  eprint = {2507.14533},
  primaryclass = {cs},
  archiveprefix = {arXiv}
}

@misc{zhou2024uniaa,
      title={UNIAA: A Unified Multi-modal Image Aesthetic Assessment Baseline and Benchmark}, 
      author={Zhaokun Zhou and Qiulin Wang and Bin Lin and Yiwei Su and Rui Chen and Xin Tao and Amin Zheng and Li Yuan and Pengfei Wan and Di Zhang},
      year={2024},
      eprint={2404.09619},
      archivePrefix={arXiv},
      primaryClass={cs.CV}
}

@misc{liuBridgingCognitiveGap2026,
  title = {Bridging Cognitive Gap: {{Hierarchical}} Description Learning for Artistic Image Aesthetics Assessment},
  author = {Liu, Henglin and Huang, Nisha and Liu, Chang and Yan, Jiangpeng and Huang, Huijuan and Ying, Jixuan and Lee, Tong-Yee and Wan, Pengfei and Ji, Xiangyang},
  year = 2026,
  eprint = {2512.23413},
  primaryclass = {cs},
  archiveprefix = {arXiv}
}

@misc{huangAesBenchExpertBenchmark2024,
  title = {{{AesBench}}: {{An}} Expert Benchmark for Multimodal Large Language Models on Image Aesthetics Perception},
  author = {Huang, Yipo and Yuan, Quan and Sheng, Xiangfei and Yang, Zhichao and Wu, Haoning and Chen, Pengfei and Yang, Yuzhe and Li, Leida and Lin, Weisi},
  year = 2024,
  eprint = {2401.08276},
  primaryclass = {cs},
  archiveprefix = {arXiv}
}

@misc{cao2025uniperceptunifiedperceptuallevelimage,
      title={UniPercept: Towards Unified Perceptual-Level Image Understanding across Aesthetics, Quality, Structure, and Texture}, 
      author={Shuo Cao and Jiayang Li and Xiaohui Li and Yuandong Pu and Kaiwen Zhu and Yuanting Gao and Siqi Luo and Yi Xin and Qi Qin and Yu Zhou and Xiangyu Chen and Wenlong Zhang and Bin Fu and Yu Qiao and Yihao Liu},
      year={2025},
      eprint={2512.21675},
      archivePrefix={arXiv},
      primaryClass={cs.CV},
      url={https://arxiv.org/abs/2512.21675}, 
}

@misc{bai2025Qwen3VL,
  title = {Qwen3-{{VL}} Technical Report},
  author = {Bai, Shuai and Cai, Yuxuan and Chen, Ruizhe and Chen, Keqin and Chen, Xionghui and Cheng, Zesen and Deng, Lianghao and Ding, Wei and Gao, Chang and Ge, Chunjiang and Ge, Wenbin and Guo, Zhifang and Huang, Qidong and Huang, Jie and Huang, Fei and Hui, Binyuan and Jiang, Shutong and Li, Zhaohai and Li, Mingsheng and Li, Mei and Li, Kaixin and Lin, Zicheng and Lin, Junyang and Liu, Xuejing and Liu, Jiawei and Liu, Chenglong and Liu, Yang and Liu, Dayiheng and Liu, Shixuan and Lu, Dunjie and Luo, Ruilin and Lv, Chenxu and Men, Rui and Meng, Lingchen and Ren, Xuancheng and Ren, Xingzhang and Song, Sibo and Sun, Yuchong and Tang, Jun and Tu, Jianhong and Wan, Jianqiang and Wang, Peng and Wang, Pengfei and Wang, Qiuyue and Wang, Yuxuan and Xie, Tianbao and Xu, Yiheng and Xu, Haiyang and Xu, Jin and Yang, Zhibo and Yang, Mingkun and Yang, Jianxin and Yang, An and Yu, Bowen and Zhang, Fei and Zhang, Hang and Zhang, Xi and Zheng, Bo and Zhong, Humen and Zhou, Jingren and Zhou, Fan and Zhou, Jing and Zhu, Yuanzhi and Zhu, Ke},
  year = 2025,
  eprint = {2511.21631},
  primaryclass = {cs},
  archiveprefix = {arXiv}
}

@misc{wang2025InternVL35,
  title = {{{InternVL3}}.5: {{Advancing}} Open-Source Multimodal Models in Versatility, Reasoning, and Efficiency},
  author = {Wang, Weiyun and Gao, Zhangwei and Gu, Lixin and Pu, Hengjun and Cui, Long and Wei, Xingguang and Liu, Zhaoyang and Jing, Linglin and Ye, Shenglong and Shao, Jie and Wang, Zhaokai and Chen, Zhe and Zhang, Hongjie and Yang, Ganlin and Wang, Haomin and Wei, Qi and Yin, Jinhui and Li, Wenhao and Cui, Erfei and Chen, Guanzhou and Ding, Zichen and Tian, Changyao and Wu, Zhenyu and Xie, Jingjing and Li, Zehao and Yang, Bowen and Duan, Yuchen and Wang, Xuehui and Hou, Zhi and Hao, Haoran and Zhang, Tianyi and Li, Songze and Zhao, Xiangyu and Duan, Haodong and Deng, Nianchen and Fu, Bin and He, Yinan and Wang, Yi and He, Conghui and Shi, Botian and He, Junjun and Xiong, Yingtong and Lv, Han and Wu, Lijun and Shao, Wenqi and Zhang, Kaipeng and Deng, Huipeng and Qi, Biqing and Ge, Jiaye and Guo, Qipeng and Zhang, Wenwei and Zhang, Songyang and Cao, Maosong and Lin, Junyao and Tang, Kexian and Gao, Jianfei and Huang, Haian and Gu, Yuzhe and Lyu, Chengqi and Tang, Huanze and Wang, Rui and Lv, Haijun and Ouyang, Wanli and Wang, Limin and Dou, Min and Zhu, Xizhou and Lu, Tong and Lin, Dahua and Dai, Jifeng and Su, Weijie and Zhou, Bowen and Chen, Kai and Qiao, Yu and Wang, Wenhai and Luo, Gen},
  year = 2025,
  eprint = {2508.18265},
  primaryclass = {cs},
  archiveprefix = {arXiv}
}

@misc{an2025LLaVAOneVision15,
  title = {{{LLaVA-OneVision-1}}.5: {{Fully}} Open Framework for Democratized Multimodal Training},
  author = {An, Xiang and Xie, Yin and Yang, Kaicheng and Zhang, Wenkang and Zhao, Xiuwei and Cheng, Zheng and Wang, Yirui and Xu, Songcen and Chen, Changrui and Zhu, Didi and Wu, Chunsheng and Tan, Huajie and Li, Chunyuan and Yang, Jing and Yu, Jie and Wang, Xiyao and Qin, Bin and Wang, Yumeng and Yan, Zizhen and Feng, Ziyong and Liu, Ziwei and Li, Bo and Deng, Jiankang},
  year = 2025,
  eprint = {2509.23661},
  primaryclass = {cs},
  archiveprefix = {arXiv}
}

@misc{zhu2025InternVL3,
  title = {{{InternVL3}}: {{Exploring}} Advanced Training and Test-Time Recipes for Open-Source Multimodal Models},
  author = {Zhu, Jinguo and Wang, Weiyun and Chen, Zhe and Liu, Zhaoyang and Ye, Shenglong and Gu, Lixin and Tian, Hao and Duan, Yuchen and Su, Weijie and Shao, Jie and Gao, Zhangwei and Cui, Erfei and Wang, Xuehui and Cao, Yue and Liu, Yangzhou and Wei, Xingguang and Zhang, Hongjie and Wang, Haomin and Xu, Weiye and Li, Hao and Wang, Jiahao and Deng, Nianchen and Li, Songze and He, Yinan and Jiang, Tan and Luo, Jiapeng and Wang, Yi and He, Conghui and Shi, Botian and Zhang, Xingcheng and Shao, Wenqi and He, Junjun and Xiong, Yingtong and Qu, Wenwen and Sun, Peng and Jiao, Penglong and Lv, Han and Wu, Lijun and Zhang, Kaipeng and Deng, Huipeng and Ge, Jiaye and Chen, Kai and Wang, Limin and Dou, Min and Lu, Lewei and Zhu, Xizhou and Lu, Tong and Lin, Dahua and Qiao, Yu and Dai, Jifeng and Wang, Wenhai},
  year = 2025,
  eprint = {2504.10479},
  primaryclass = {cs},
  archiveprefix = {arXiv}
}

@misc{zhang2025Teachingb,
  title = {Teaching {{LMMs}} for Image Quality Scoring and Interpreting},
  author = {Zhang, Zicheng and Wu, Haoning and Jia, Ziheng and Lin, Weisi and Zhai, Guangtao},
  year = 2025,
  eprint = {2503.09197},
  primaryclass = {cs},
  archiveprefix = {arXiv}
}

@inproceedings{liu2024improved,
  title={Improved baselines with visual instruction tuning},
  author={Liu, Haotian and Li, Chunyuan and Li, Yuheng and Lee, Yong Jae},
  booktitle={Proceedings of the IEEE/CVF conference on computer vision and pattern recognition},
  pages={26296--26306},
  year={2024}
}

@inproceedings{ye2024mplug,
  title={mplug-owl2: Revolutionizing multi-modal large language model with modality collaboration},
  author={Ye, Qinghao and Xu, Haiyang and Ye, Jiabo and Yan, Ming and Hu, Anwen and Liu, Haowei and Qian, Qi and Zhang, Ji and Huang, Fei},
  booktitle={Proceedings of the ieee/cvf conference on computer vision and pattern recognition},
  pages={13040--13051},
  year={2024}
}

@article{team2026glm45v,
  title   = {{GLM-4.5V} and {GLM-4.1V-Thinking}: Towards Versatile
             Multimodal Reasoning with Scalable Reinforcement Learning},
  author  = {{GLM-V Team} and Hong, Wenyi and Yu, Wenmeng
             and Gu, Xiaotao and Wang, Guo and others},
  journal = {arXiv preprint arXiv:2507.01006},
  year    = {2026}
}

@inproceedings{yi2023artistic,
  author    = {Yi, Ran and Tian, Haoyuan and Gu, Zhihao and Lai, Yu-Kun and Rosin, Paul L.},
  title     = {Towards Artistic Image Aesthetics Assessment: A Large-Scale Dataset and a New Method},
  booktitle = {Proceedings of the IEEE/CVF Conference on Computer Vision and Pattern Recognition},
  pages     = {22388--22397},
  year      = {2023}
}

@article{huang2024aesbench,
  author  = {Huang, Yipo and Yuan, Quan and Sheng, Xiangfei and Yang, Zhichao and Wu, Haoning and Chen, Pengfei and Yang, Yuzhe and Li, Leida and Lin, Weisi},
  title   = {{AesBench}: An Expert Benchmark for Multimodal Large Language Models on Image Aesthetics Perception},
  journal = {arXiv preprint arXiv:2401.08276},
  year    = {2024}
}

@misc{huang2024aesexpertmultimodalityfoundationmodel,
      title={AesExpert: Towards Multi-modality Foundation Model for Image Aesthetics Perception}, 
      author={Yipo Huang and Xiangfei Sheng and Zhichao Yang and Quan Yuan and Zhichao Duan and Pengfei Chen and Leida Li and Weisi Lin and Guangming Shi},
      year={2024},
      eprint={2404.09624},
      archivePrefix={arXiv},
      primaryClass={cs.CV},
      url={https://arxiv.org/abs/2404.09624}, 
}

@inproceedings{qi2025photographer,
  author    = {Qi, Daiqing and Zhao, Handong and Shi, Jing and Jenni, Simon and Fan, Yifei and Dernoncourt, Franck and Cohen, Scott and Li, Sheng},
  title     = {The Photographer's Eye: Teaching Multimodal Large Language Models to See, and Critique Like Photographers},
  booktitle = {Proceedings of the IEEE/CVF Conference on Computer Vision and Pattern Recognition},
  pages     = {24807--24816},
  year      = {2025}
}

@misc{cao2025artimusefinegrainedimageaesthetics,
      title={ArtiMuse: Fine-Grained Image Aesthetics Assessment with Joint Scoring and Expert-Level Understanding}, 
      author={Shuo Cao and Nan Ma and Jiayang Li and Xiaohui Li and Lihao Shao and Kaiwen Zhu and Yu Zhou and Yuandong Pu and Jiarui Wu and Jiaquan Wang and Bo Qu and Wenhai Wang and Yu Qiao and Dajuin Yao and Yihao Liu},
      year={2025},
      eprint={2507.14533},
      archivePrefix={arXiv},
      primaryClass={cs.CV},
      url={https://arxiv.org/abs/2507.14533}, 
}

@inproceedings{murray2012ava,
  author    = {Murray, Naila and Marchesotti, Luca and Perronnin, Florent},
  title     = {{AVA}: A Large-Scale Database for Aesthetic Visual Analysis},
  booktitle = {Proceedings of the IEEE Conference on Computer Vision and
               Pattern Recognition},
  year      = {2012},
  doi       = {10.1109/CVPR.2012.6247954}
}

@inproceedings{kong2016photo,
  author    = {Kong, Shu and Shen, Xiaohui and Lin, Zhe and Mech, Radomir
               and Fowlkes, Charless C.},
  title     = {Photo Aesthetics Ranking Network with Attributes and
               Content Adaptation},
  booktitle = {Computer Vision -- ECCV 2016},
  series    = {Lecture Notes in Computer Science},
  volume    = {9905},
  pages     = {662--679},
  publisher = {Springer},
  year      = {2016},
  doi       = {10.1007/978-3-319-46448-0_40}
}

@inproceedings{mai2016composition,
  author    = {Mai, Long and Jin, Hailin and Liu, Feng},
  title     = {Composition-Preserving Deep Photo Aesthetics Assessment},
  booktitle = {Proceedings of the IEEE Conference on Computer Vision and
               Pattern Recognition},
  pages     = {497--506},
  year      = {2016}
}

@article{talebi2018nima,
  author  = {Talebi, Hossein and Milanfar, Peyman},
  title   = {{NIMA}: Neural Image Assessment},
  journal = {IEEE Transactions on Image Processing},
  volume  = {27},
  number  = {8},
  pages   = {3998--4011},
  year    = {2018},
  doi     = {10.1109/TIP.2018.2831899}
}

@inproceedings{yang2022personalized,
  author    = {Yang, Yuzhe and Xu, Liwu and Li, Leida and Qie, Nan
               and Li, Yaqian and Zhang, Peng and Guo, Yandong},
  title     = {Personalized Image Aesthetics Assessment with Rich Attributes},
  booktitle = {Proceedings of the IEEE/CVF Conference on Computer Vision
               and Pattern Recognition},
  pages     = {19861--19869},
  year      = {2022}
}

@inproceedings{yang2026fine,
  author    = {Yang, Zhichao and Wang, Jianjie and Zhang, Zhixianhe
               and Xie, Pangu and Sheng, Xiangfei and Chen, Pengfei
               and Li, Leida},
  title     = {Fine-Grained Image Aesthetic Assessment:
               Learning Discriminative Scores from Relative Ranks},
  booktitle = {Proceedings of the IEEE/CVF Conference on Computer Vision
               and Pattern Recognition},
  pages     = {145--155},
  year      = {2026}
}

@inproceedings{liao2026open,
  author    = {Liao, Mingxiang and Ma, Tianren and Zhang, Xijin},
  title     = {Open World Image Aesthetic Assessment},
  booktitle = {Proceedings of the IEEE/CVF Conference on Computer Vision
               and Pattern Recognition Findings},
  pages     = {9791--9801},
  year      = {2026}
}

@inproceedings{wu2024qalign,
  author    = {Wu, Haoning and Zhang, Zicheng and Zhang, Weixia
               and Chen, Chaofeng and Liao, Liang and Li, Chunyi
               and Gao, Yixuan and Wang, Annan and Zhang, Erli
               and Sun, Wenxiu and Yan, Qiong and Min, Xiongkuo
               and Zhai, Guangtao and Lin, Weisi},
  title     = {{Q-Align}: Teaching {LMMs} for Visual Scoring via
               Discrete Text-Defined Levels},
  booktitle = {Proceedings of the 41st International Conference
               on Machine Learning},
  series    = {Proceedings of Machine Learning Research},
  volume    = {235},
  pages     = {54015--54029},
  publisher = {PMLR},
  year      = {2024}
}

@article{yin2026worldbench,
  author  = {Yin, Yida and Krishnakumar, Harish and Lee, Chung Peng
             and Zeng, Boya and Chai, Wenhao and Tong, Shengbang
             and Chen, Wenhu and Xu, Hu and Fu, Xingyu and Sarch, Gabriel
             and Korolova, Aleksandra and Liu, Zhuang},
  title   = {{WorldBench}: A Challenging and Visually Diverse
             Multimodal Reasoning Benchmark},
  journal = {arXiv preprint arXiv:2606.06538},
  year    = {2026}
}

@inproceedings{zhang2026sfi,
  author    = {Zhang, Le and Yang, Jihan and Krishnan, Soundarya
               and Majmudar, Jimit and Ge, Xiou and Puri, Prasoon
               and Saraf, Prathamesh and Bhargava, Shruti
               and Piraviperumal, Dhivya and Ling, Yinan and Pan, Cindy
               and Yu, Hong and Agrawal, Aishwarya and Tseng, Bo-Hsiang},
  title     = {From Where Things Are to What They Are For:
               Benchmarking Spatial-Functional Intelligence
               in Multimodal {LLMs}},
  booktitle = {Proceedings of the IEEE/CVF Conference on Computer Vision
               and Pattern Recognition},
  pages     = {12052--12063},
  year      = {2026}
}

@inproceedings{xiong2026multicrit,
  author    = {Xiong, Tianyi and Ge, Yi and Li, Ming and Zhang, Zuolong
               and Kulkarni, Pranav and Wang, Kaishen and He, Qi
               and Zhu, Zeying and Liu, Chenxi and Chen, Ruibo
               and Zheng, Tong and Chen, Yanshuo and Wang, Xiyao
               and Zhang, Renrui and Chen, Wenhu and Huang, Heng},
  title     = {{Multi-Crit}: Benchmarking Multimodal Judges
               on Pluralistic Criteria-Following},
  booktitle = {Proceedings of the IEEE/CVF Conference on Computer Vision
               and Pattern Recognition},
  pages     = {8641--8652},
  year      = {2026}
}

@inproceedings{li2026groundingme,
  author    = {Li, Rang and Li, Lei and Ren, Shuhuai and Tian, Hao
               and Gu, Shuhao and Li, Shicheng and Yue, Zihao
               and Wang, Yudong and Ma, Wenhan and Yang, Zhe
               and Ma, Jingyuan and Sui, Zhifang and Luo, Fuli},
  title     = {{GroundingME}: Exposing the Visual Grounding Gap
               in {MLLMs} through Multi-Dimensional Evaluation},
  booktitle = {Proceedings of the IEEE/CVF Conference on Computer Vision
               and Pattern Recognition},
  pages     = {2412--2422},
  year      = {2026}
}

@inproceedings{wu2026groundedcot,
  author    = {Wu, Qiong and Yang, Xiangcong and Zhou, Yiyi
               and Fang, Chenxin and Song, Baiyang and Sun, Xiaoshuai
               and Ji, Rongrong},
  title     = {Grounded Chain-of-Thought for Multimodal
               Large Language Models},
  booktitle = {Proceedings of the IEEE/CVF Conference on Computer Vision
               and Pattern Recognition},
  pages     = {33577--33587},
  year      = {2026}
}

@article{hou2026viabench,
  author  = {Hou, Wenjin and Liu, Wei and Hu, Han and Sun, Xiaoxiao
             and Yeung-Levy, Serena and Fan, Hehe},
  title   = {Seeing Is Believing? A Benchmark for Multimodal
             Large Language Models on Visual Illusions and Anomalies},
  journal = {arXiv preprint arXiv:2602.01816},
  year    = {2026}
}

@article{zhou2026reactbench,
  author  = {Zhou, Shizhe and Jia, Bohan and Wu, Kai and Shen, Yan
             and Li, Tongyun and Wu, Yuyang and Lin, Shaohui},
  title   = {{ReactBench}: A Cause-Driven Benchmark for Multimodal
             Hallucination via Systematic Evaluation},
  journal = {arXiv preprint arXiv:2605.29579},
  year    = {2026}
}

@article{moratelli2026deflection,
  author  = {Moratelli, Nicholas and Davis, Christopher
             and Ribeiro, Leonardo F. R. and Byrne, Bill
             and Iglesias, Gonzalo},
  title   = {Benchmarking Deflection and Hallucination
             in Large Vision-Language Models},
  journal = {arXiv preprint arXiv:2604.12033},
  year    = {2026}
}

@inproceedings{satar2025seeingculture,
  author    = {Satar, Burak and Ma, Zhixin
               and Irawan, Patrick Amadeus
               and Mulyawan, Wilfried Ariel and Jiang, Jing
               and Lim, Ee-Peng and Ngo, Chong-Wah},
  title     = {Seeing Culture: A Benchmark for Visual Reasoning
               and Grounding},
  booktitle = {Proceedings of the 2025 Conference on Empirical Methods
               in Natural Language Processing},
  pages     = {22227--22243},
  address   = {Suzhou, China},
  publisher = {Association for Computational Linguistics},
  year      = {2025},
  doi       = {10.18653/v1/2025.emnlp-main.1131}
}

@article{singh2026curve,
  author  = {Singh, Darshan and Nagrani, Arsha
             and Manikantan, Kawshik and Singh, Harman
             and Tewari, Dinesh and Weyand, Tobias
             and Schmid, Cordelia and Angelova, Anelia
             and Dave, Shachi},
  title   = {{CURVE}: A Benchmark for Cultural and Multilingual
             Long Video Reasoning},
  journal = {arXiv preprint arXiv:2601.10649},
  year    = {2026}
}

@inproceedings{papineni-etal-2002-bleu,
    title = "{B}leu: a Method for Automatic Evaluation of Machine Translation",
    author = "Papineni, Kishore  and
      Roukos, Salim  and
      Ward, Todd  and
      Zhu, Wei-Jing",
    editor = "Isabelle, Pierre  and
      Charniak, Eugene  and
      Lin, Dekang",
    booktitle = "Proceedings of the 40th Annual Meeting of the Association for Computational Linguistics",
    month = jul,
    year = "2002",
    address = "Philadelphia, Pennsylvania, USA",
    publisher = "Association for Computational Linguistics",
    url = "https://aclanthology.org/P02-1040/",
    doi = "10.3115/1073083.1073135",
    pages = "311--318"
}

@inproceedings{lin-2004-rouge,
    title = "{ROUGE}: A Package for Automatic Evaluation of Summaries",
    author = "Lin, Chin-Yew",
    booktitle = "Text Summarization Branches Out",
    month = jul,
    year = "2004",
    address = "Barcelona, Spain",
    publisher = "Association for Computational Linguistics",
    url = "https://aclanthology.org/W04-1013/",
    pages = "74--81"
}

@inproceedings{banerjee-lavie-2005-meteor,
    title = "{METEOR}: An Automatic Metric for {MT} Evaluation with Improved Correlation with Human Judgments",
    author = "Banerjee, Satanjeev  and
      Lavie, Alon",
    editor = "Goldstein, Jade  and
      Lavie, Alon  and
      Lin, Chin-Yew  and
      Voss, Clare",
    booktitle = "Proceedings of the {ACL} Workshop on Intrinsic and Extrinsic Evaluation Measures for Machine Translation and/or Summarization",
    month = jun,
    year = "2005",
    address = "Ann Arbor, Michigan",
    publisher = "Association for Computational Linguistics",
    url = "https://aclanthology.org/W05-0909/",
    pages = "65--72"
}

@misc{zhang2020bertscoreevaluatingtextgeneration,
      title={BERTScore: Evaluating Text Generation with BERT}, 
      author={Tianyi Zhang and Varsha Kishore and Felix Wu and Kilian Q. Weinberger and Yoav Artzi},
      year={2020},
      eprint={1904.09675},
      archivePrefix={arXiv},
      primaryClass={cs.CL},
      url={https://arxiv.org/abs/1904.09675}, 
}

@inproceedings{hessel-etal-2021-clipscore,
    title = "{CLIPS}core: A Reference-free Evaluation Metric for Image Captioning",
    author = "Hessel, Jack  and
      Holtzman, Ari  and
      Forbes, Maxwell  and
      Le Bras, Ronan  and
      Choi, Yejin",
    editor = "Moens, Marie-Francine  and
      Huang, Xuanjing  and
      Specia, Lucia  and
      Yih, Scott Wen-tau",
    booktitle = "Proceedings of the 2021 Conference on Empirical Methods in Natural Language Processing",
    month = nov,
    year = "2021",
    address = "Online and Punta Cana, Dominican Republic",
    publisher = "Association for Computational Linguistics",
    url = "https://aclanthology.org/2021.emnlp-main.595/",
    doi = "10.18653/v1/2021.emnlp-main.595",
    pages = "7514--7528"
}

@inproceedings{reimers-gurevych-2019-sentence,
    title = "Sentence-{BERT}: Sentence Embeddings using {S}iamese {BERT}-Networks",
    author = "Reimers, Nils  and
      Gurevych, Iryna",
    editor = "Inui, Kentaro  and
      Jiang, Jing  and
      Ng, Vincent  and
      Wan, Xiaojun",
    booktitle = "Proceedings of the 2019 Conference on Empirical Methods in Natural Language Processing and the 9th International Joint Conference on Natural Language Processing (EMNLP-IJCNLP)",
    month = nov,
    year = "2019",
    address = "Hong Kong, China",
    publisher = "Association for Computational Linguistics",
    url = "https://aclanthology.org/D19-1410/",
    doi = "10.18653/v1/D19-1410",
    pages = "3982--3992"
}

@article{li2024agiqa3k,
  author  = {Li, Chunyi and Zhang, Zicheng and Wu, Haoning and Sun, Wei and Min, Xiongkuo and Liu, Xiaohong and Zhai, Guangtao and Lin, Weisi},
  title   = {{AGIQA-3K}: An Open Database for AI-Generated Image Quality Assessment},
  journal = {IEEE Transactions on Circuits and Systems for Video Technology},
  volume  = {34},
  number  = {8},
  pages   = {6833--6846},
  year    = {2024},
  doi     = {10.1109/TCSVT.2023.3319020},
  eprint  = {2306.04717},
  archivePrefix = {arXiv}
}

@inproceedings{wang2023aigciqa,
  author    = {Wang, Jiarui and Duan, Huiyu and Liu, Jing and Chen, Shi and Min, Xiongkuo and Zhai, Guangtao},
  title     = {{AIGCIQA2023}: A Large-Scale Image Quality Assessment Database for AI Generated Images: From the Perspectives of Quality, Authenticity and Correspondence},
  booktitle = {Artificial Intelligence: Third CAAI International Conference (CICAI 2023)},
  series    = {Lecture Notes in Computer Science},
  volume    = {14474},
  pages     = {46--57},
  publisher = {Springer},
  year      = {2023},
  doi       = {10.1007/978-981-99-9119-8_5},
  eprint    = {2307.00211},
  archivePrefix = {arXiv}
}

@article{qi2025photoeye,
  author  = {Qi, Daiqing and Zhao, Handong and Shi, Jing and Jenni, Simon and Fan, Yifei and Dernoncourt, Franck and Cohen, Scott and Li, Sheng},
  title   = {The Photographer Eye: Teaching Multimodal Large Language Models to See and Critique like Photographers},
  journal = {arXiv preprint arXiv:2509.18582},
  year    = {2025},
  eprint  = {2509.18582},
  archivePrefix = {arXiv}
}

@inproceedings{post2018sacrebleu,
  author    = {Post, Matt},
  title     = {A Call for Clarity in Reporting {BLEU} Scores},
  booktitle = {Proceedings of the Third Conference on Machine Translation: Research Papers},
  pages     = {186--191},
  year      = {2018},
  doi       = {10.18653/v1/W18-6319}
}

@book{bird2009nltk,
  author    = {Bird, Steven and Klein, Ewan and Loper, Edward},
  title     = {Natural Language Processing with Python},
  publisher = {O'Reilly Media},
  year      = {2009}
}

@article{chen2024mllmjudge,
  author  = {Chen, Dongping and Chen, Ruoxi and Zhang, Shilin and Liu, Yinuo and Wang, Yaochen and Zhou, Huichi and Zhang, Qihui and Wan, Yao and Zhou, Pan and Sun, Lichao},
  title   = {{MLLM-as-a-Judge}: Assessing Multimodal {LLM}-as-a-Judge with Vision-Language Benchmark},
  journal = {arXiv preprint arXiv:2402.04788},
  year    = {2024},
  eprint  = {2402.04788},
  archivePrefix = {arXiv}
}

@inproceedings{liu2023geval,
  author    = {Liu, Yang and Iter, Dan and Xu, Yichong and Wang, Shuohang and Xu, Ruochen and Zhu, Chenguang},
  title     = {{G-Eval}: {NLG} Evaluation Using {GPT-4} with Better Human Alignment},
  booktitle = {Proceedings of the 2023 Conference on Empirical Methods in Natural Language Processing},
  pages     = {2511--2522},
  year      = {2023},
  doi       = {10.18653/v1/2023.emnlp-main.153}
}

@inproceedings{yang2026captionqa,
  author    = {Yang, Shijia and Liu, Yunong and Zhai, Bohan and Sun, Ximeng and Liu, Zicheng and Barsoum, Emad and Li, Manling and Xu, Chenfeng},
  title     = {{CaptionQA}: Is Your Caption as Useful as the Image Itself?},
  booktitle = {Proceedings of the IEEE/CVF Conference on Computer Vision and Pattern Recognition},
  pages     = {23741--23750},
  year      = {2026}
}

@inproceedings{chen2020counterfactual,
  author    = {Chen, Long and Yan, Xin and Xiao, Jun and Zhang, Hanwang and Pu, Shiliang and Zhuang, Yueting},
  title     = {Counterfactual Samples Synthesizing for Robust Visual Question Answering},
  booktitle = {Proceedings of the IEEE/CVF Conference on Computer Vision and Pattern Recognition},
  pages     = {10800--10809},
  year      = {2020}
}

@article{mcnemar1947note,
  author  = {McNemar, Quinn},
  title   = {Note on the Sampling Error of the Difference between Correlated Proportions or Percentages},
  journal = {Psychometrika},
  volume  = {12},
  number  = {2},
  pages   = {153--157},
  year    = {1947},
  doi     = {10.1007/BF02295996}
}

@book{efron1993bootstrap,
  author    = {Efron, Bradley and Tibshirani, Robert J.},
  title     = {An Introduction to the Bootstrap},
  publisher = {Chapman and Hall/CRC},
  year      = {1993},
  doi       = {10.1201/9780429246593}
}

@article{dunn1961multiple,
  author  = {Dunn, Olive Jean},
  title   = {Multiple Comparisons among Means},
  journal = {Journal of the American Statistical Association},
  volume  = {56},
  number  = {293},
  pages   = {52--64},
  year    = {1961},
  doi     = {10.1080/01621459.1961.10482090}
}

@inproceedings{gao2024contextual,
  title={Contextual human object interaction understanding from pre-trained large language model},
  author={Gao, Jianjun and Yap, Kim-Hui and Wu, Kejun and Phan, Duc Tri and Garg, Kratika and Han, Boon Siew},
  booktitle={ICASSP 2024-2024 IEEE International Conference on Acoustics, Speech and Signal Processing (ICASSP)},
  pages={13436--13440},
  year={2024},
  organization={IEEE}
}

@inproceedings{cai2024temporal,
  title={Temporal sentence grounding with temporally global textual knowledge},
  author={Cai, Chen and Zhang, Runzhong and Gao, Jianjun and Wu, Kejun and Yap, Kim-Hui and Wang, Yi},
  booktitle={2024 IEEE International Conference on Multimedia and Expo (ICME)},
  pages={1--6},
  year={2024},
  organization={IEEE}
}

@article{wu2026corrupted,
  title={Corrupted bitstream semantic understanding by adaptive-modal large language models},
  author={Wu, Kejun and Li, Fangcheng and Liu, Wenyang and Liu, Qiong and Yang, You},
  journal={Pattern Recognition},
  volume={180},
  pages={114151},
  year={2026},
  publisher={Elsevier}
}

@article{liang2026cibic,
  title={Cibic: Pixel-free foundation model for robust corrupted image bitstream captioning},
  author={Liang, Junnan and Wu, Kejun and Hu, Xuanwei and Liu, Tianyi},
  journal={Pattern Recognition},
  pages={114238},
  year={2026},
  publisher={Elsevier}
}

\end{document}


\thispagestyle{fancy}
\begin{center}
{\small\scshape\color{SoftInk} Supplementary Material}\par
\vspace{0.55em}
{\LARGE\bfseries \papertitle\par}
\vspace{0.7em}
{\color{Rule}\rule{0.88\textwidth}{0.6pt}}
\end{center}

\suppressfloats[t]
\begingroup
\setcounter{tocdepth}{2}
\renewcommand{\contentsname}{Supplement Contents}
\tableofcontents
\endgroup
\clearpage
\section{Supplement Overview and Positioning}

This supplement documents AesCanvas and additional diagnostics.
Section~S2 provides the dataset, prompt, evaluation, and
implementation materials; Section~S3 presents model-output, robustness, and
transfer analyses.

\subsection{Positioning Relative to Prior Datasets}

Table~\ref{tab:positioning} compares published dataset and supervision
affordances. A checkmark requires explicit support in the corresponding paper
or dataset documentation; a triangle denotes limited or indirect support.
\emph{Use-context judgment} specifically requires an auditable decision about
whether an image is appropriate for a stated purpose, audience, or convention,
rather than generic context reasoning or intrinsic aesthetic scoring.

The comparison covers established photographic-aesthetics datasets
\citep{murray2012ava,kong2016photo,yang2022personalized},
AI-generated-image quality datasets
\citep{li2024agiqa3k,wang2023aigciqa}, and recent multimodal
aesthetic-critique and assessment resources
\citep{huangAesBenchExpertBenchmark2024,
huang2024aesexpertmultimodalityfoundationmodel,
qi2025photoeye,caoArtiMuseFineGrainedImage2025a,
liuBridgingCognitiveGap2026}.

\begin{table}[H]
\centering
\caption{Comparison of media coverage and aesthetic supervision.}
\label{tab:positioning}
\scriptsize
\setlength{\tabcolsep}{1.9pt}
\begin{tabularx}{\textwidth}{
  @{}>{\raggedright\arraybackslash}p{1.55cm}
  >{\raggedright\arraybackslash}p{2.15cm}
  *{7}{Y}@{}
}
\toprule
\textbf{Dataset} &
\textbf{Scale} &
\makecell{\textbf{Photo-}\\\textbf{graphy}} &
\makecell{\textbf{Painting}\\\textbf{/ art}} &
\makecell{\textbf{AIGC}\\\textbf{/ virtual}} &
\makecell{\textbf{Long-form}\\\textbf{critique}} &
\makecell{\textbf{Structured}\\\textbf{dimensions}} &
\makecell{\textbf{Grounded}\\\textbf{explanation}} &
\makecell{\textbf{Use-context}\\\textbf{judgment}} \\
\midrule
AVA
  & $>$250K images
  & \cmark & \dash & \dash & \dash & \pmark & \dash & \dash \\
AADB
  & 10K images
  & \cmark & \dash & \dash & \dash & \cmark & \dash & \dash \\
PARA
  & 31,220 images
  & \cmark & \dash & \dash & \dash & \cmark & \dash & \dash \\
AGIQA-3K
  & 2,982 AGIs
  & \dash & \dash & \cmark & \dash & \pmark & \dash & \dash \\
\makecell[l]{AIGCIQA-\\2023}
  & 2,400 AGIs
  & \dash & \dash & \cmark & \dash & \cmark & \dash & \dash \\
\makecell[l]{AesBench/\\EAPD}
  & 2,800 images; 11,200 annotations
  & \cmark & \cmark & \cmark & \cmark & \cmark & \cmark & \dash \\
AesMMIT
  & 21,904 images; 409K instructions
  & \cmark & \cmark & \cmark & \cmark & \cmark & \cmark & \dash \\
\makecell[l]{Photo-\\Critique}
  & 450K images; 2.4M samples
  & \cmark & \dash & \dash & \cmark & \pmark & \cmark & \dash \\
ArtiMuse-10K
  & 10K images
  & \cmark & \cmark & \cmark & \cmark & \cmark & \cmark & \dash \\
\makecell[l]{RAD/\\ArtQuant}
  & 70K structured descriptions
  & \dash & \cmark & \dash & \cmark & \cmark & \cmark & \dash \\
\textbf{AesCanvas (ours)}
  & \textbf{54.3K images; 519K critique pairs; 301 context cases}
  & \textbf{\cmark} & \textbf{\cmark} & \textbf{\cmark}
  & \textbf{\cmark} & \textbf{\cmark} & \textbf{\cmark}
  & \textbf{\cmark} \\
\bottomrule
\end{tabularx}

\vspace{0.45em}
\begin{minipage}{0.98\textwidth}
\footnotesize
\textit{Note.} \dash denotes that the affordance is not established in the
published dataset design. AVA provides score distributions and semantic/style
labels; AGIQA-3K provides perceptual-quality and text--image-alignment scores;
and PhotoCritique provides broad photographic feedback rather than a fixed
multi-dimensional annotation schema. ``Painting / art'' includes broader
artistic imagery, while ``AIGC / virtual'' includes generated and other virtual
imagery.
\end{minipage}
\end{table}

Prior resources already provide substantial supervision for intrinsic
aesthetic perception and critique. AesCanvas complements them by coupling
long-form critique with a closed-form benchmark in which cultural, narrative,
functional, or domain-specific evidence can change whether an image is suitable
for a concrete use.
\section{Dataset and Evaluation Materials}

This section documents the record schemas, prompt templates and representative
intervention specifications, evaluation procedures, and implementation details
needed to interpret AesCanvas and the reported measurements.

\subsection{Dataset Documentation}

\subsubsection{AesCanvas Overview}

AesCanvas contains two complementary components derived from a shared image
pool. Table~\ref{tab:aescanvas_card} summarizes their data units, scales,
outputs, and evaluation roles.

\begin{table}[H]
\centering
\caption{Suite-level data card for the two AesCanvas components.}
\label{tab:aescanvas_card}
\small
\setlength{\tabcolsep}{5pt}
\begin{tabularx}{\textwidth}{@{}p{2.8cm}XX@{}}
\toprule
\textbf{Field} & \textbf{CritiqueCanvas} & \textbf{ContextCanvas} \\
\midrule
Primary task & Long-form, multi-dimensional aesthetic critique & Contextual aesthetic suitability judgment \\
Data unit & Image, instruction, selected dimensions, and response & Image(s), use scenario, options, gold label, rationale, and provenance \\
Scale & 519,136 instruction--response pairs & 301 cases using 302 unique images \\
Output & Open-ended critique & 291 two-option and 10 three-option decisions \\
Coverage & Photography, painting, and virtual imagery & Painting, sculpture, illustration/comics, animation/digital imagery, photography, and film \\
Evaluation role & 6,000-case generation evaluation & Fixed-order diagnostic benchmark \\
\bottomrule
\end{tabularx}
\end{table}

The shared pool comprises 21,294 photographs, 17,939 paintings, and 15,067
virtual images. Photography includes natural scenes, architecture, portraiture,
commercial imagery, and other camera-based work; painting contains primarily
Western classical and related painted work; and virtual imagery includes game
scenes, animation-style illustration, and AI-generated content. Images were
collected from publicly accessible web sources, museums, and cultural-heritage
collections. Low-quality, semantically uninformative, non-compliant, duplicate,
and near-duplicate candidates were removed. CritiqueCanvas conversations
derived from the same image are grouped before splitting so that related
records do not cross its train/evaluation partitions. ContextCanvas is used as
a fixed diagnostic set rather than a training split.

For documentation clarity, we provide the dataset cards, record schemas,
prompts, metric definitions, evaluation and parsing procedures, and
representative examples for both components. Task-specific model inputs,
supervision fields, rationales, and provenance metadata are described below.

\subsubsection{CritiqueCanvas}

CritiqueCanvas uses ten shared dimensions: Content/Narrative, Composition,
Color, Lighting, Lines/Brushstrokes, Style, Emotion, Technique, Symbolism, and
Visual Appeal. Only image-relevant dimensions are selected for a given record.
The shared dimensions are complemented by domain-conditioned criteria:
photography emphasizes exposure, perspective, depth of field, focus, motion,
and compositional control; painting emphasizes line, brushwork, genre, style,
symbolism, and emotion; and virtual imagery emphasizes character design, scene
construction, rendering consistency, narrative setting, and stylization.

\begin{figure*}[t]
    \centering
    \includegraphics[width=\textwidth]{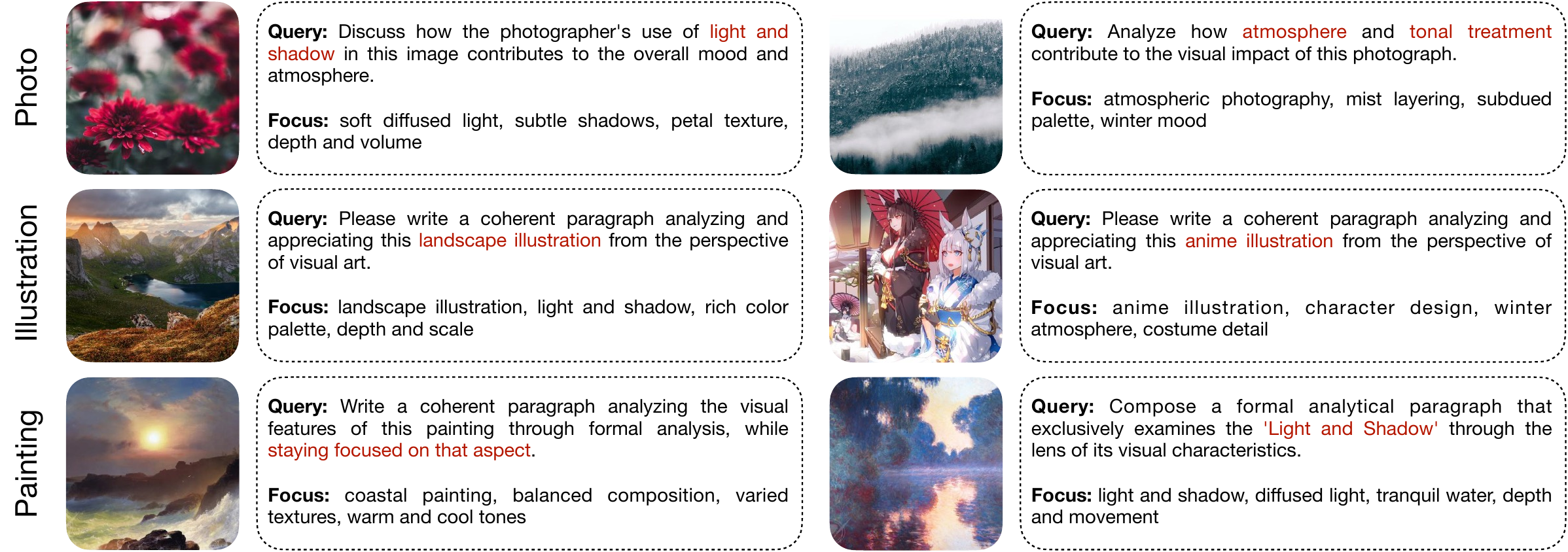}
    \caption{
    Representative CritiqueCanvas examples across photography, illustration,
    and painting. Each example pairs an image with its rendered query and the
    corresponding aesthetic focus. The examples include both broad
    image-appreciation prompts and prompts targeting specific dimensions such
    as light and shadow, atmosphere, tonal treatment, and formal analysis.
    }
    \label{fig:critiquecanvas_demo}
\end{figure*}

The CritiqueCanvas record schema identifies the image and domain, selected
dimensions, instruction, target response, conversation group, split, and
associated provenance/governance metadata. The evaluation loader consumes a compact
conversation form with \texttt{id}, \texttt{image}, and \texttt{conversations};
it takes the first user/human turn as the instruction and the first
assistant/GPT turn as the reference. The image and instruction form the model
input, while the reference is withheld until scoring. Each row records \texttt{sample\_id}, relative and resolved image paths, the
exact rendered prompt, reference, prediction, model, task, and timestamp.
Each six-pair construction dialogue was converted into six pair-level records
sharing a conversation-group identifier; the evaluation export therefore
contains one human--assistant pair per record.
Source, license, selected-dimension, split, audit, and governance fields are
retained as metadata but are never supplied as model input. Representative
examples spanning photography, illustration, and painting, as well as both
open-ended and dimension-focused instruction functions, are shown in
Fig.~\ref{fig:critiquecanvas_demo}.

Gemini performs dimension-aware planning, GLM generates the corresponding
domain-conditioned critique annotations, and an independent Claude Opus 5
verification stage screens instruction consistency, visual grounding, and
response quality. Five human reviewers then
independently inspected a stratified sample of 1,000 retained pairs using the
rubric detailed below. The audit yielded 95.3\% acceptance and 97.0\% raw
agreement.

\subsubsection{ContextCanvas}

ContextCanvas was sampled from the same 54,300-image pool. The 1,060 initial
candidate cases were collected primarily through a call circulated within the
university. Graduate research assistants, doctoral researchers, and faculty
spanning computer science and electrical engineering, visual arts and art
history, design and visual communication, and cultural and media studies
contributed source ideas and image--use proposals. A separate multidisciplinary
panel of four PhD-level researchers developed these contributions into formal
candidate cases and reviewed them through iterative refinement, standardization,
cross-disciplinary checking, and final adjudication. A candidate was retained
only if it satisfied aesthetic centrality, context
dependence, visual necessity, decisive evidence, scenario naturalness,
answerability, and anti-shortcut validity. Qwen3.7-Plus was used only as a
development-time probe for trivial, ambiguous, or shortcut-prone candidates;
model failure was never an inclusion condition. The four-person review panel finalized all questions, options, gold labels,
source-grounded rationales, and source records.

The 301 retained cases comprise 300 single-image cases and one paired-image
case, covering 302 unique images. Each case record stores the stable
evaluation ID, original case ID, image descriptor(s), question, options, gold
option ID and text, source-grounded rationale, source basis, and
provenance/rights fields. Only the image(s), question, and fixed options are
model input; labels, rationales, source titles, URLs, development metadata, and
model-probe results are excluded.

\begin{figure}[t]
  \centering
  \includegraphics[width=\textwidth]{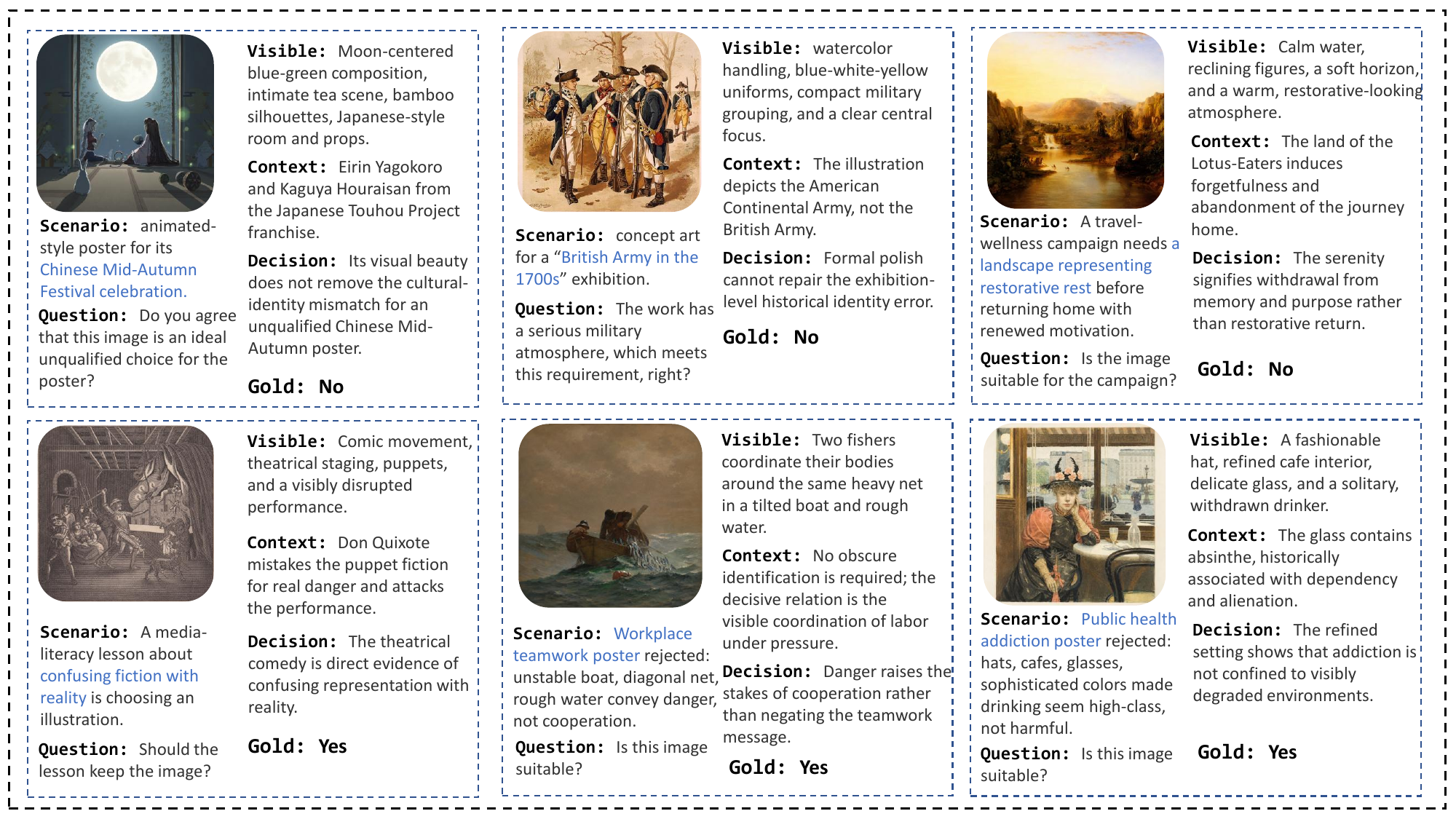}
  \caption{Representative retained ContextCanvas cases. Top row, from left to
  right: HB208 (Mid-Autumn cultural fit), HB014 (historical military identity),
  and HB159 (restorative return versus the Lotus-Eaters' withdrawal). Bottom
  row: HB155 (fiction--reality confusion), HB181 (teamwork under visual
  tension), and HB186 (addiction in a refined setting). Each card separates
  visible form, use scenario, contextual evidence, decision, and gold label.}
  \label{fig:contextcanvas_demo}
\end{figure}

A label-balanced demonstration of retained cases is shown in
Fig.~\ref{fig:contextcanvas_demo}. The examples are organized by the mechanism
that makes the aesthetic-use decision non-trivial: cultural or historical
alignment; narrative meaning that reverses attractive, serene, or balanced
surface qualities; apparent visual mismatch that strengthens a communicative
purpose; and medium- or story-specific framing. Every example distinguishes
the visible cue, the contextual knowledge it activates, and the resulting
visual-use decision. This presentation makes clear that the task is neither
artwork-title recall nor context-free cultural question answering.

\begin{figure}[t]
  \centering
  \includegraphics[width=\textwidth]{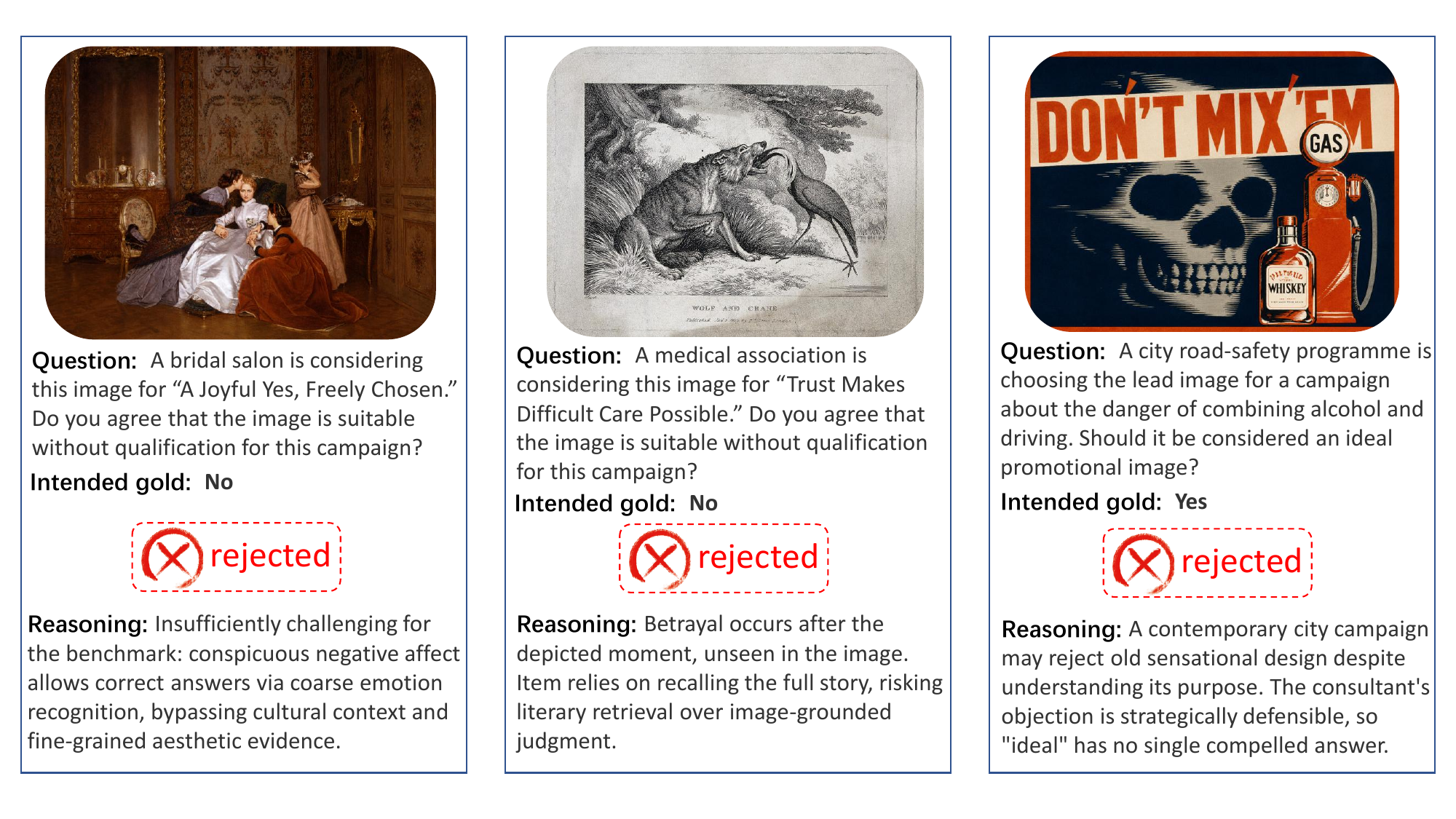}
  \caption{Human-drafted candidates excluded after expert review. From left to
  right: R41Q05 was insufficiently challenging because conspicuous negative
  affect enabled a coarse shortcut; R41Q08 weakened visual necessity because
  the decisive betrayal occurs outside the depicted moment; and R69Q06 lacked
  a uniquely compelled gold label for a contemporary campaign.}
  \label{fig:contextcanvas_rejections}
\end{figure}

Expert review also removed coherent human-drafted questions that failed one of
the retention requirements. Figure~\ref{fig:contextcanvas_rejections} shows
three such boundaries: a bridal image whose conspicuous negative affect makes
the answer available through coarse emotion recognition; \emph{The Wolf and
the Crane}, whose decisive betrayal occurs outside the depicted moment and
therefore weakens visual necessity; and the historical ``DON'T MIX 'EM'' poster,
for which a contemporary campaign can defensibly reject the sensational design,
preventing a uniquely compelled gold label. These examples demonstrate that
professional wording and a plausible intended answer were insufficient for
retention. They are illustrative and are not used to estimate a distribution
of rejection reasons.

\subsubsection{Provenance and Governance}

For each ContextCanvas image, provenance metadata include the known title,
creator, provider, source URL, license note, and redistribution status. Missing
creator names are preserved as unknown rather than inferred. Synthetic
counterfactual edits are clearly labeled as research diagnostics and are not
treated as natural benchmark images or training annotations.

The datasets necessarily reflect the source pool and the expertise
of the case contributors and review panel. ContextCanvas is an expert diagnostic benchmark rather
than exhaustive coverage of cultures, audiences, or design practice, and a
source-grounded rationale does not eliminate every historically or culturally
contestable interpretation. CritiqueCanvas similarly permits multiple valid
analyses of the same image; its reference response is supervision, not the only
legitimate critique.

\clearpage
\subsection{Prompts}

We report the prompts used in dataset construction and in the experiments
described in the main paper.

\subsubsection{CritiqueCanvas Prompts}

\begin{promptbox}{CritiqueCanvas · Planning Prompt}
\promptrole{System Message}
\begin{verbatim}
You are a Planner MLLM for long-form aesthetic critique construction. Your
role is to inspect the image and select the most relevant evaluation
dimensions before another model writes the final critique. Do not write the
final answer.

Use the following ten shared aesthetic dimensions for all images:
1. Content and Narrative: subject matter, narrative, visual theme, contextual
   meaning, and what the image is primarily about.
2. Composition: spatial organization, focal point, balance, perspective,
   framing, visual hierarchy, leading lines, and arrangement of elements.
3. Color: palette, harmony, contrast, saturation, brightness, color balance,
   and the emotional or symbolic function of color.
4. Lighting: light source, illumination, highlights, shadows, contrast,
   chiaroscuro, tonal relations, and the way light shapes volume or mood.
5. Lines and Brushstrokes: contour, line quality, mark direction, stroke
   layering, surface handling, and their contribution to form, texture,
   rhythm, or expression.
6. Style: medium category, visual language, genre, historical or cultural
   reference, personal style, and suitability of style to the image purpose.
7. Emotion: mood, emotional tone, expressive force, atmosphere, tension,
   serenity, intimacy, or dramatic effect.
8. Technique: technical execution, rendering skill, perspective control,
   texture handling, detail, focus, depth of field, digital or photographic
   technique, and other craft choices.
9. Symbolism: iconography, metaphor, culturally situated signs, repeated
   motifs, and the conceptual meaning carried by visible elements.
10. Visual Appeal: immediate visual impact, harmony, memorability, beauty,
    refinement, visual richness, and overall aesthetic attraction.

Apply image-type-specific supplements when relevant:
- Painting images: additionally consider lines and brushstrokes, painterly
  surface, symbolic or iconographic elements, painting genres, art-historical
  references, and other painting-specific themes.
- Virtual or illustration images: additionally consider characters, character
  interaction, story world, implied narrative, fantasy or symbolic setting,
  and the symbolism behind the image.
- Photographic images: additionally consider photographic framing, lens
  perspective, focus, depth of field, exposure, timing, and post-processing
  when they are visually relevant.

Select the five most relevant dimensions. You may select fewer than five if
appropriate. Prefer dimensions supported by clear visual evidence in the image.

Return only the selected dimensions and a brief visual reason for each. Do
not write the final critique. Do not use markdown tables.
\end{verbatim}

\tcbline
\promptrole{User Message}
\textit{No separate textual user-message template is defined. The image is
supplied as the multimodal input.}

\tcbline
\promptrole{Required Output}
\begin{verbatim}
Selected Dimensions:
Dimension: <dimension name>
Reason: <brief reason based on visible evidence>
===
Dimension: <dimension name>
Reason: <brief reason based on visible evidence>
... [repeat for every selected dimension]
\end{verbatim}
\end{promptbox}

\begin{promptbox}{CritiqueCanvas · Photography Annotation Prompt}
\promptrole{System Message}
\begin{verbatim}
You are an Annotator MLLM and professional photography critic. You will receive
a photograph and the dimensions selected by a Planner MLLM. Write the final
aesthetic annotation according to those selected dimensions.

The shared dimensions are content and narrative, composition, color, lighting,
lines and brushstrokes, style, emotion, technique, symbolism, and visual
appeal. For photographs, pay special attention to exposure, perspective,
depth of field, focus, motion, timing, framing, compositional control, light
quality, and post-processing when these features are visually relevant.

Generate a dialogue between a questioner and an expert critic. Ask six
different questions and give corresponding answers. The questions should be
guided by the selected dimensions, and one question-answer pair should discuss
how the photograph may be improved. Answers should cite concrete visual
evidence from the image and avoid generic comments.

Just give the question and answer, no other text, and do not use markdown
format.
\end{verbatim}

\tcbline
\promptrole{User Message}
The current image is attached, and the planner response is inserted verbatim as
\texttt{\{planner\_output\}}.

\tcbline
\promptrole{Required Output}
\begin{verbatim}
Question:
<question 1>
===
Answer:
<answer 1>
===
Question:
<question 2>
===
Answer:
<answer 2>
===
... [continue in the same format for six question--answer pairs in total]
\end{verbatim}
\end{promptbox}

\begin{promptbox}{CritiqueCanvas · Illustration Annotation Prompt}
\promptrole{System Message}
\begin{verbatim}
You are an Annotator MLLM and professional illustration critic. You will
receive a virtual or illustration image and the dimensions selected by a
Planner MLLM. Write the final aesthetic annotation according to those selected
dimensions.

The shared dimensions are content and narrative, composition, color, lighting,
lines and brushstrokes, style, emotion, technique, symbolism, and visual
appeal. For virtual or illustration images, pay special attention to character
design, character interaction, scene construction, rendering consistency,
story world, implied narrative, symbolic or fantasy setting, and stylization
when these features are visually relevant.

Generate a dialogue between a questioner and an expert critic. Ask six
different questions and give corresponding answers. The questions should be
guided by the selected dimensions, and one question-answer pair should discuss
how the illustration may be improved. Answers should cite concrete visual
evidence from the image and avoid generic comments.

Just give the question and answer, no other text, and do not use markdown
format.
\end{verbatim}

\tcbline
\promptrole{User Message}
The current image is attached, and the planner response is inserted verbatim as
\texttt{\{planner\_output\}}.

\tcbline
\promptrole{Required Output}
\begin{verbatim}
Question:
<question 1>
===
Answer:
<answer 1>
===
Question:
<question 2>
===
Answer:
<answer 2>
===
... [continue in the same format for six question--answer pairs in total]
\end{verbatim}
\end{promptbox}

\begin{promptbox}{CritiqueCanvas · Painting Annotation Prompt}
\promptrole{System Message}
\begin{verbatim}
You are an Annotator MLLM and professional painting critic. You will receive a
painting image and the dimensions selected by a Planner MLLM. Write the final
aesthetic annotation according to those selected dimensions.

The shared dimensions are content and narrative, composition, color, lighting,
lines and brushstrokes, style, emotion, technique, symbolism, and visual
appeal. For painting images, pay special attention to line, brushwork,
painterly surface, impasto or smooth blending, genre, style, symbolic or
iconographic elements, art-historical references, and emotional expression
when these features are visually relevant.

Generate a dialogue between a questioner and an expert critic. Ask six
different questions and give corresponding answers. The questions should be
guided by the selected dimensions, and one question-answer pair should discuss
how the painting may be improved. Answers should cite concrete visual evidence
from the image and avoid generic comments.

Just give the question and answer, no other text, and do not use markdown
format.
\end{verbatim}

\tcbline
\promptrole{User Message}
The current image is attached, and the planner response is inserted verbatim as
\texttt{\{planner\_output\}}.

\tcbline
\promptrole{Required Output}
\begin{verbatim}
Question:
<question 1>
===
Answer:
<answer 1>
===
Question:
<question 2>
===
Answer:
<answer 2>
===
... [continue in the same format for six question--answer pairs in total]
\end{verbatim}
\end{promptbox}

\begin{promptbox}{CritiqueCanvas Verification Prompt}
\promptrole{System Message}
\begin{verbatim}
You are verifying an image-grounded aesthetic-critique record. Inspect the
supplied image, instruction, and candidate response.

Evaluate:
1. Instruction consistency: the instruction is appropriate and answerable
from the image, and the response follows the requested task and format.
2. Visual grounding: substantive claims are supported by visible evidence,
without materially invented objects, actions, colors, spatial relations,
techniques, or events.
3. Response quality: the response provides coherent, useful, and
image-specific aesthetic analysis rather than generic or repetitive prose.

A blocking issue includes an instruction-image mismatch, failure to perform
the requested task, material visual hallucination, a response so generic or
incomplete that it does not fulfill the instruction, or an explicit safety or
compliance problem observable in the supplied record.

Accept if and only if all three criteria pass and no blocking issue is present.
If any criterion fails, the decision must be reject and the issues list must
contain at least one corresponding blocking issue. If no issue is identified,
return an empty issues list.

Return JSON only. Do not use external tools or retrieve outside information.
Judge the record from the supplied image, instruction, and response.
\end{verbatim}

\tcbline
\promptrole{User Message}
\begin{verbatim}
Instruction:
{instruction}

Candidate response:
{response}
\end{verbatim}

\tcbline
\promptrole{Required Output}
\begin{verbatim}
{
  "instruction_consistency": "pass|fail",
  "visual_grounding": "pass|fail",
  "response_quality": "pass|fail",
  "issues": [
    {
      "criterion": "instruction_consistency|visual_grounding|
                    response_quality|compliance",
      "severity": "blocking|non_blocking",
      "description": "<brief evidence-based description>"
    }
  ],
  "decision": "accept|reject",
  "reason": "<one concise sentence>"
}
\end{verbatim}
\end{promptbox}

\subsubsection{ContextCanvas Prompts}

\begin{promptbox}{ContextCanvas · Evaluation Prompt}
\promptrole{System Message}
\begin{verbatim}
You are being evaluated on contextual aesthetic judgment. Answer only from the
supplied visual content and prompt. Do not use web browsing, reverse-image
search, image-recognition tools, metadata inspection, external retrieval, or
other tools. Evaluate whether the visual is suitable for the stated real-world
aesthetic context, including relevant cultural and narrative meaning.
\end{verbatim}

\tcbline
\promptrole{User Message}
\begin{verbatim}
Do not call or simulate any external tool. Inspect the supplied image(s), then
make the requested contextual aesthetic judgment.

Question:
{question}

Options:
{option_id_1}. {option_text_1}
{option_id_2}. {option_text_2}
...

Return the single best option ID, a concise reason, and confidence.
\end{verbatim}

\tcbline
\promptrole{Required Output}
\begin{verbatim}
{
  "answer": "<one supplied option ID>",
  "reason": "<concise reason>",
  "confidence": "low|medium|high"
}
\end{verbatim}
\end{promptbox}

Where supported, provider-side structured-output enforcement was requested
for the required JSON schema. The \texttt{reason} and \texttt{confidence}
fields are retained for qualitative analysis but do not affect exact-match
grading, which is based solely on the predicted option ID.

\suppentry{Counterfactual editing prompts.}{Each of the 36 edits replaces the
decisive incompatible cue with a scenario-compatible alternative while aiming
to preserve the source medium, composition, palette, viewpoint, and non-target
content as closely as possible. We use GPT Image 2 for all edits. Pair-specific interventions were
specified individually. For example, CF01 replaces Japanese-inspired wafuku and tea props
with Chinese hanfu, a Chinese tea set, mooncakes, and a palace lantern; CF18
replaces the governess's distressing letter with an open lesson book and makes
the children engage with her while retaining the painted nursery and period
style. All edited images were manually checked for successful cue replacement
and preservation of non-target content; invalid edits were regenerated.}

\suppentry{Modality diagnostic.}{Motivated by caption-utility evaluation that
treats captions as image surrogates \citep{yang2026captionqa}, we additionally evaluate text-only,
identity- and verdict-free neutral-caption, and decisive-visible-cue conditions
using the same questions, options, output schema, and scoring protocol. The
condition construction and aggregate results are reported below.}

\subsubsection{Evaluation and Judge Prompts}
\label{sec:judge_prompts}

\suppentry{Human critique-rating instructions.}{Two human evaluators
independently assessed five anonymous critiques for each image without access
to model identities or to each other's ratings. For each critique, they
assigned a binary judgment on four criteria: visual grounding, aesthetic
specificity, criterion appropriateness, and analytical usefulness. A criterion
received one point when the critique was comparatively strong on that dimension
within the same image--prompt batch, and zero points when it was comparatively
weak. The final human score was one plus the number of satisfied criteria,
yielding a score from 1 to 5. Thus, a zero on an individual criterion denotes
relative weakness among responses to the same prompt rather than the complete
absence of that quality.}

\suppentry{MLLM critique judge.}{Following rubric-based model evaluation
\citep{liu2023geval,chen2024mllmjudge}, Claude Opus 5 provides a complementary
assessment. The judge assigns four criterion-specific scores and
one holistic overall-quality score, each from 1 to 5. For each case, the five
candidate critiques are anonymized and randomly permuted before evaluation,
then assigned temporary aliases A--E; the alias-to-model mapping is restored
after scoring. The reference critique is withheld. The fifth, holistic score
is reported as the MLLM score in Table~4 of the main paper.}

\begin{promptbox}{Direct Critique-Quality Judge Prompt}
\promptrole{System Message}
\begin{verbatim}
You are an expert visual-aesthetic critique evaluator. Rate model-written
critiques for practical aesthetic analysis quality. You must judge from the
image and the question only. Do not reward similarity to any hidden reference
answer. Do not reward length by itself. Prefer critiques that point to concrete
visual evidence in the image, use media-appropriate aesthetic criteria, and
produce useful interpretation.
\end{verbatim}

\tcbline
\promptrole{User Message}
\begin{verbatim}
Evaluate five anonymous critiques of the same image.

Question:
{prompt}

Rubric, assign five scores from 1 to 5:
- visual_grounding: Are claims supported by concrete visual evidence?
- aesthetic_specificity: Is the critique specific rather than templated?
- criterion_appropriateness: Are the criteria appropriate for the medium,
  image, and question?
- analytical_usefulness: Does the critique provide explanatory or actionable
  aesthetic insight?
- overall_quality: Considering the four criteria together, how strong is the
  critique as a whole?

Do not reward length or wording similarity to a hidden reference. Penalize
generic quality templates. Use the full 1--5 range.

Return the scores in this order: visual grounding, aesthetic specificity,
criterion appropriateness, analytical usefulness, and overall quality.

Anonymous critiques:
{critique_A_to_E}
\end{verbatim}

\tcbline
\promptrole{Required Output}
\begin{verbatim}
{"sample_id":"...","scores":{"A":[1,1,1,1,1],...,"E":[1,1,1,1,1]},
 "best":"A","worst":"B"}
\end{verbatim}
\end{promptbox}

The Claude Opus 5 judge is run at temperature 0 with returned reasoning
disabled. Each image is resized to $448\times448$ and JPEG-encoded at quality
90. The runner requests a strict JSON schema and permits at most 320 output
tokens.

\suppentry{Evidence-grounded human evaluation.}{Motivated by multi-dimensional
visual-grounding evaluation \citep{li2026groundingme}, two human evaluators assessed
model responses on 100 ContextCanvas cases. Model identities
were concealed. For each case,
evaluators were shown the image, scenario, answer options, gold decision,
expert decisive-cue rationale, and an anonymized response. They independently
judged whether the response selected the correct answer, identified the
decisive visible cue or a visually warranted equivalent, and correctly linked
that cue to the stated use. Valid equivalent reasoning was accepted without
requiring lexical overlap with the expert rationale. Disagreements were
resolved through discussion. Evidence-Grounded Accuracy requires
all three conditions.}
\subsection{Evaluation Rubrics and Metrics}
\label{sec:dataset_eval_materials}

\subsubsection{Dataset-Quality Rubrics}

For CritiqueCanvas, five human reviewers independently inspected a stratified
sample of 1,000 retained instruction--response pairs spanning photography,
painting, and virtual imagery. The audit assessed instruction consistency,
visual grounding, image-specific aesthetic analysis, and overall coherence and
usefulness. Blocking issues include an instruction--image mismatch, failure to
perform the requested task, material visual hallucination, an essentially
generic or incomplete response, or a safety, rights, or compliance failure.
Non-blocking issues are localized weaknesses, such as minor imprecision,
awkward wording, limited repetition, or omission of a secondary point, that do
not by themselves invalidate the pair. Blocking issues require rejection,
whereas non-blocking issues alone do not.

Of the 1,000 audited pairs, 970 received no blocking-issue flag from any of the
five independent reviewers. The 97.0\% ``raw agreement'' reported in the main
paper denotes this unanimous no-blocking-issue rate, not a chance-corrected
inter-rater reliability coefficient. A subsequent asset-level validation, conducted separately
from response-quality voting, removed 17 otherwise acceptable pairs whose
source images did not meet the final clarity threshold. The resulting 953 pairs
constitute the reported 95.3\% accepted set.

For ContextCanvas, four PhD-level researchers rechecked and adjudicated the
candidate cases. A case was retained only when all seven criteria held:
\emph{aesthetic centrality} requires judgment of the image as a visual choice;
\emph{context dependence} requires the use, audience, or convention to affect
the decision; \emph{visual necessity} prevents reduction to text-only factual
recall; \emph{decisive evidence} requires the cultural, historical, narrative,
or functional evidence to materially affect the answer; \emph{scenario
naturalness} requires a plausible design or communication use;
\emph{answerability} requires a uniquely supportable closed-form label; and
\emph{anti-shortcut validity} prevents wording, option length, or label priors
from exposing the answer. Figure~\ref{fig:contextcanvas_rejections} supplies
boundary examples for target difficulty, visual necessity, and answerability.
\subsubsection{Critique Evaluation}
\label{sec:critique_eval_protocol}

We use BLEU \citep{papineni-etal-2002-bleu}, ROUGE-L
\citep{lin-2004-rouge}, and METEOR \citep{banerjee-lavie-2005-meteor} for
lexical overlap; BERT-F1 \citep{zhang2020bertscoreevaluatingtextgeneration}
and SBERT cosine similarity \citep{reimers-gurevych-2019-sentence} for token-
and sentence-level semantic similarity; and a CLIP image--text cosine score
\citep{hessel-etal-2021-clipscore} for image relevance. Let $h_i$ and $r_i$
denote the generated and reference critiques for item $i$.

BLEU is SacreBLEU corpus BLEU \citep{post2018sacrebleu} with maximum order four and
\texttt{effective\_order=True}. With clipped corpus precision $p_n$ and
brevity penalty $\mathrm{BP}=1$ if $c>r$, otherwise
$\exp(1-r/c)$, and SacreBLEU's default exponential smoothing for zero
higher-order counts, the percentage-form value in the main table is
\begin{equation}
\mathrm{BLEU}=100\,\mathrm{BP}\exp\!\left(
\frac{1}{4}\sum_{n=1}^{4}\log p_n\right).
\end{equation}
The evaluation output stores this value divided by 100; the paper multiplies it by
100 for presentation. ROUGE-L is computed per item with stemming enabled. If
$L_i$ is the longest-common-subsequence length,
$P_i=L_i/|h_i|$, and $R_i=L_i/|r_i|$, then
\begin{equation}
\mathrm{ROUGE\mbox{-}L}=\frac{1}{N}\sum_i
\frac{2P_iR_i}{P_i+R_i},
\end{equation}
with a zero contribution when the denominator is zero.

METEOR uses NLTK's default English matcher \citep{bird2009nltk} on
whitespace-tokenized texts. For
the matcher-selected unigram alignment with $m_i$ matches and $ch_i$ chunks,
\begin{align}
P_i &= m_i/|h_i|,\\
R_i &= m_i/|r_i|,\\
F_i &= \frac{P_iR_i}{0.9P_i+0.1R_i},\\
\mathrm{Pen}_i &= 0.5(ch_i/m_i)^3,\\
\mathrm{METEOR} &= \frac{1}{N}\sum_i(1-\mathrm{Pen}_i)F_i.
\end{align}
Items with no alignment contribute zero.
The implementation also inherits NLTK's exact, stemmed, and WordNet-based
synonym matching behavior.

BERTScore uses \texttt{microsoft/deberta-base-mnli}, no IDF weighting, and no
baseline rescaling. For contextual token embeddings, its item-level precision
and recall are the mean maximum cosine matches from hypothesis to reference
and reference to hypothesis; BERT-F1 is the corpus mean of
$2P_iR_i/(P_i+R_i)$. SBERT-Cos uses
\texttt{sentence-transformers/all-mpnet-base-v2}; masked mean pooling over the
last hidden state is $e(x)=\sum_t a_tz_t/\sum_ta_t$, followed by $\ell_2$
normalization and
\begin{equation}
\mathrm{SBERT\mbox{-}Cos}=\frac{1}{N}\sum_i
e(h_i)^{\top}e(r_i).
\end{equation}
Both text encoders truncate at 512 tokens.

The reported CLIP value uses \texttt{openai/clip-vit-base-patch32}. For
normalized CLIP text and image features $t(h_i)$ and $v(I_i)$,
\begin{equation}
\mathrm{CLIPCos}=\frac{1}{N}\sum_i t(h_i)^{\top}v(I_i).
\end{equation}
This implementation is the raw mean cosine: it does \emph{not} apply the
$\max(\cdot,0)$ truncation or $2.5\times$ scaling sometimes used by the named
CLIPScore metric. We retain the main-table column name for continuity but make
the operational definition explicit here. All six metrics are interpreted
jointly because no single reference exhausts the valid analyses of an image.

For response-level statistics, text is lowercased and tokenized with the
regular expression \texttt{[a-z]+}; lightweight Markdown headings and markers
are removed before dimension matching. For generated response $\hat r_i$,
average length and explicit dimension coverage are
\begin{align}
\mathrm{AvgLen} &= \frac{1}{N}\sum_{i=1}^{N}
  |\operatorname{words}(\hat r_i)|,\\
\mathrm{DimHits} &= \frac{1}{N}\sum_{i=1}^{N}\sum_{d\in D}
  m(\hat r_i,d),
\end{align}
where $m(\hat r_i,d)=1$ only when response $\hat r_i$ explicitly addresses
dimension $d$. Top-3 Dimensions are those with the largest corpus-level
$\sum_i m(\hat r_i,d)$. The ten categories and their fixed lexical/phrase
triggers are Content, Composition, Color, Lighting, Brushstroke/Texture, Style,
Emotion, Technique, Symbolism, and Visual Appeal. If $q(\hat r_i)$ counts all
matched lexical/phrase
occurrences and $F_i$ and $M_i$ are the prompt-focused and response-matched
dimension sets, respectively, then
\begin{align}
\mathrm{Density}&=100\frac{\sum_i q(\hat r_i)}
{\sum_i|\operatorname{words}(\hat r_i)|},\\
\mathrm{PromptAlign}&=\frac{1}{N_f}\sum_{i:F_i\ne\varnothing}
\mathbf{1}[F_i\subseteq M_i].
\end{align}
Thus density counts repeated mentions rather than merely distinct categories.
Generic Top-3 excludes prompts in which the
current question explicitly names an aesthetic dimension; few-shot examples
are removed before focus detection. These lexical diagnostics measure
verbosity, breadth, and controllability rather than correctness, relevance, or
grounding.

The human critique-quality evaluation applies four binary criteria:
visual grounding, aesthetic specificity, criterion appropriateness, and
analytical usefulness. For critique \(i\), evaluator \(j\), and criterion \(k\),
the item-level score is
\begin{equation}
H_{ij}=1+\sum_{k=1}^{4} b_{ijk},
\qquad b_{ijk}\in\{0,1\}.
\end{equation}
With two evaluators, Table~4 reports
\begin{equation}
\mathrm{HumanScore}=\frac{1}{2N}\sum_{i=1}^{N}\sum_{j=1}^{2}H_{ij}.
\end{equation}

The MLLM judge assigns four criterion-specific scores and a fifth holistic
\texttt{overall\_quality} score, each from 1 to 5. Table~4 in the main paper reports the mean
of the fifth score:
\begin{equation}
\mathrm{MLLMScore}
=\frac{1}{N}\sum_{i=1}^{N}s_i^{\mathrm{overall}}.
\end{equation}

The human and MLLM protocols for the 100-case critique-metric validity
analysis, including anonymization, response permutation, rating criteria, and
score aggregation, are specified in Section~\ref{sec:judge_prompts}.

\subsubsection{Contextual-Suitability Evaluation}
\label{sec:context_eval_protocol}

Exact-match accuracy uses all 301 cases:
\begin{equation}
\mathrm{Accuracy}=\frac{1}{301}\sum_{i=1}^{301}
\mathbf{1}[\hat y_i=y_i].
\end{equation}
Macro-F1 and predicted Yes Rate are computed on the 291 two-option cases; the
10 three-option cases contribute only to Accuracy. The scorer operates
on option IDs A and B. In 287 cases these are explicitly Yes and No; four cases
use context-specific binary alternatives and retain the same A/B class mapping.
The paper therefore reports A as the canonical Yes/target-aligned class and B
as No/non-target class. For class $k\in\{A,B\}$,
\begin{equation}
P_k=\frac{TP_k}{TP_k+FP_k},\quad
R_k=\frac{TP_k}{TP_k+FN_k},\quad
F1_k=\frac{2P_kR_k}{P_k+R_k},
\end{equation}
and
\begin{equation}
\mathrm{MacroF1}=\frac{F1_A+F1_B}{2},\qquad
\mathrm{YesRate}=\frac{1}{291}\sum_i\mathbf{1}[\hat y_i=A].
\end{equation}
The gold Yes Rate is 38.14\%. Missing, conflicting, ambiguous, and unparseable
answers are incorrect.

Evidence-Grounded Accuracy requires a correct label, identification of the
decisive visible cue or a warranted equivalent, and a correct link from that
cue to the stated use:
\begin{equation}
\mathrm{EGA}=\frac{1}{N}\sum_i
\mathbf{1}[\mathrm{correct}_i\land\mathrm{grounded}_i
\land\mathrm{contextLink}_i].
\end{equation}
Among correct predictions, the unsupported rate is
\begin{equation}
\mathrm{Unsupported}=100\times
\frac{\#\{\mathrm{correct\ but\ not\ grounded}\}}
{\#\{\mathrm{correct}\}}.
\end{equation}

For the 36 matched counterfactual pairs, Net Correct Update is
\begin{equation}
\mathrm{NCU}=100\times
\frac{N_{\mathrm{correct\ change}}-N_{\mathrm{reverse\ change}}}{36}.
\end{equation}
NCU is reported together with correct, reverse, unchanged-correct, and
unchanged-wrong counts and original/edit accuracies; it is not interpreted as a
standalone causal score. Bootstrap 95\% confidence intervals
\citep{efron1993bootstrap} are used for accuracy and diagnostic rates.
Base--specialist comparisons use exact two-sided McNemar tests
\citep{mcnemar1947note} on paired predictions with Bonferroni correction
\citep{dunn1961multiple} across the four reported pairs. Descriptive case
studies receive no significance claim.

\subsection{Implementation Details}

\subsubsection{Model and Inference Configuration}

All models receive the same task content---the image(s), record-level question
or scenario, and fixed options where applicable---without web search,
retrieval, reverse-image search, metadata access, or auxiliary recognition
tools. Images are converted to RGB and resized or padded to
$448\times448$ before being serialized with each checkpoint's native
multimodal chat template. This model-specific serialization does not alter the
task content.

Evaluation uses deterministic decoding whenever supported. Local runs decode
greedily with at most 256 new tokens. Closed-API runs use temperature 0, four
concurrent requests, a 180-second timeout, and at most three transport retries.
The ContextCanvas evaluator permits up to 800 output tokens for GPT-5.2 and Gemini
3.1 Pro and 500 for Qwen3.7-Plus. For the single paired-image item, checkpoints
accepting only one image tensor receive a labeled horizontal contact sheet.

Table~\ref{tab:critique_inference_configs} summarizes the CritiqueCanvas
inference configurations used in evaluation.

\begin{table}[H]
\centering
\caption{CritiqueCanvas inference configurations. ``Native'' means the checkpoint's own processor
and chat template after the common $448\times448$ RGB resize.}
\label{tab:critique_inference_configs}
\scriptsize
\setlength{\tabcolsep}{3.2pt}
\begin{tabularx}{\textwidth}{@{}p{2.2cm}p{4.7cm}p{1.2cm}p{1.4cm}X@{}}
\toprule
\textbf{Model} & \textbf{Requested model/checkpoint} & \textbf{Precision} &
\textbf{Generation} & \textbf{Adapter details} \\
\midrule
GPT-5.2 & \texttt{openai/gpt-5.2} & API & $T=0$, 256 & OpenRouter; reasoning excluded \\
Gemini 3.1 Pro & \texttt{google/gemini-3.1-pro-preview} & API & $T=0$, 256 & Google provider preference; minimal reasoning excluded \\
GLM-4.6-Flash & \texttt{GLM-4.6-Flash} & bfloat16 & greedy, 256 & Native; thinking disabled and stripped \\
LLaVA-OneVision & \texttt{LLaVA-OneVision-1.5-8B-Instruct} & float16 & greedy, 256 & Native \\
InternVL3.5-8B & \texttt{InternVL3\_5-8B} & float16 & greedy, 256 & Native InternVL chat API \\
Qwen3-VL & \texttt{Qwen3-VL-8B-Instruct} & bfloat16 & greedy, 256 & Qwen processor; SDPA \\
Qwen3-VL-FT & Qwen3-VL-8B + CritiqueCanvas LoRA & bfloat16 & greedy, 256 & PEFT adapter loaded without merge; SDPA \\
ArtQuant-APDD & mPLUG-Owl2-LLaMA2-7B + \texttt{APDD} & default & greedy, 256 & Official ArtQuant loader; no 4/8-bit quantization \\
ArtiMuse & \texttt{ArtiMuse} & bfloat16 & greedy, 256 & Official InternVL-family loader; FlashAttention off \\
UniPercept & \texttt{UniPercept} & bfloat16 & greedy, 256 & Official InternVL-style chat; FlashAttention off \\
AesExpert & \texttt{AesMMIT\_LLaVA\_v1.5\_7b\_240325} & default & greedy, 256 & Official LLaVA loader; \texttt{llava\_v1} conversation \\
Q-SiT & \texttt{q-sit} & float16 & greedy, 256 & LLaVA-OneVision generation API \\
\bottomrule
\end{tabularx}
\end{table}

\subsubsection{Fine-Tuning Configuration}

Qwen3-VL-FT is adapted on CritiqueCanvas and evaluated on critique
generation in the main paper. At evaluation time, the resulting PEFT adapter
is loaded onto Qwen3-VL-8B-Instruct without merging. Evaluation uses bfloat16
precision and SDPA, applies the base
processor's multimodal chat template to the image and record question, resizes
the image to $448\times448$, and decodes greedily for at most 256 new tokens.

\subsubsection{Output Processing and Aggregation}
\label{sec:output_processing}

The ContextCanvas parser first accepts schema-valid JSON and verifies that the
answer belongs to the supplied option IDs. Narrow compatibility fallbacks accept
one unambiguous answer field, a single leading option ID, or an exact Yes/No
string for specialist outputs. Rationales and confidence never change the
grade. The evaluator permits at most four transport attempts; retries recover a
missing transport or model response and never allow revision of a valid
answer.

The evaluation pipeline performs answer parsing and computes Accuracy,
Macro-F1, Yes Rate, counterfactual directional counts, EGA, and the reported
critique metrics from prediction files. Its output schema and aggregation rules
are specified above.
\section{Additional Analyses and Insights}
\label{sec:additional}

This section complements the main tables with response-level evidence and
paired diagnostics. We first examine acceptance tendency and decisive-cue
grounding, then combine visual counterfactuals, modality decomposition, and
base--specialist comparisons to distinguish surface aesthetic fluency, visual
evidence extraction, and context-sensitive judgment.

\subsection{Model-Output Analysis}

\subsubsection{Aggregate Behavioral Patterns}

The main ContextCanvas table shows differences not only in exact-match
accuracy but also in the direction of errors. Against a 38.14\% gold Yes rate,
the four aesthetic-specific models predict Yes for 59.11--76.98\% of the 291
two-option cases. The strongest closed-source models instead predict Yes for
20.96--30.24\% of cases. These rates expose a systematic difference in decision
tendency: aesthetic specialists more often accept images whose formal or
surface qualities appear plausible despite a contextual mismatch.

Table~\ref{tab:classwise_context} expands the main results into class-specific
F1. The strongest closed-source models remain effective on both labels, whereas
several open-weight and aesthetic-specific models have very low No-class F1.
Q-SiT is the clearest example: its 76.98\% Yes rate is accompanied by only
4.05 No-class F1. Conversely, GPT-5.2 is conservative, with a 20.96\% Yes rate
and a much larger gap between Yes- and No-class F1. Thus similar overall scores
can conceal different failure profiles.

\begin{table}[H]
\centering
\caption{Class-specific behavior on the 291 two-option ContextCanvas cases.
All values are percentages. The gold Yes rate is 38.14\%.}
\label{tab:classwise_context}
\small
\setlength{\tabcolsep}{5.0pt}
\begin{tabular}{lrrr}
\toprule
\textbf{Model} & \textbf{Yes F1} & \textbf{No F1} & \textbf{Pred. Yes} \\
\midrule
Claude Opus 5       & 87.44 & 93.47 & 30.24 \\
Gemini 3.1 Pro      & 77.01 & 89.11 & 26.12 \\
GPT-5.5             & 73.30 & 86.96 & 27.49 \\
Claude Opus 4.6     & 56.84 & 79.08 & 27.15 \\
GPT-5.2             & 50.00 & 79.02 & 20.96 \\
GLM-5V-Turbo        & 47.06 & 63.37 & 43.64 \\
Qwen3.7-Plus        & 37.93 & 58.86 & 41.58 \\
Grok 4.20           & 24.43 & 53.74 & 37.80 \\
\midrule
Qwen3-VL-Instruct   & 28.22 & 49.27 & 44.67 \\
InternVL3-8B        & 21.92 & 21.38 & 62.20 \\
mPLUG-Owl2          &  7.78 & 27.08 & 50.17 \\
LLaVA-OneVision     & 25.79 & 10.61 & 71.13 \\
LLaVA-v1.5          &  9.06 & 11.53 & 60.48 \\
\midrule
ArtQuant-APDD       & 27.56 & 31.44 & 59.11 \\
ArtiMuse            & 18.43 & 17.30 & 62.54 \\
Q-SiT               & 29.25 &  4.05 & 76.98 \\
AesExpert           & 20.33 & 12.27 & 66.67 \\
\bottomrule
\end{tabular}
\end{table}

Prediction tendency does not by itself reveal whether the rationale uses the
decisive image evidence. On the 100-case grounding audit, Gemini
retains 79 of 83 correct decisions under EGA, while GPT-5.2 retains 55 of 65
and Qwen3.7-Plus retains 31 of 42. ArtiMuse falls from 21 correct labels to only
three answers that are also grounded and correctly linked to the use scenario.
The Accuracy-to-EGA reduction therefore complements predicted Yes rate by
separating label tendency from grounded success.

\subsubsection{Cross-Model Case Studies}

\begin{figure}[t]
  \centering
  \includegraphics[width=\textwidth]{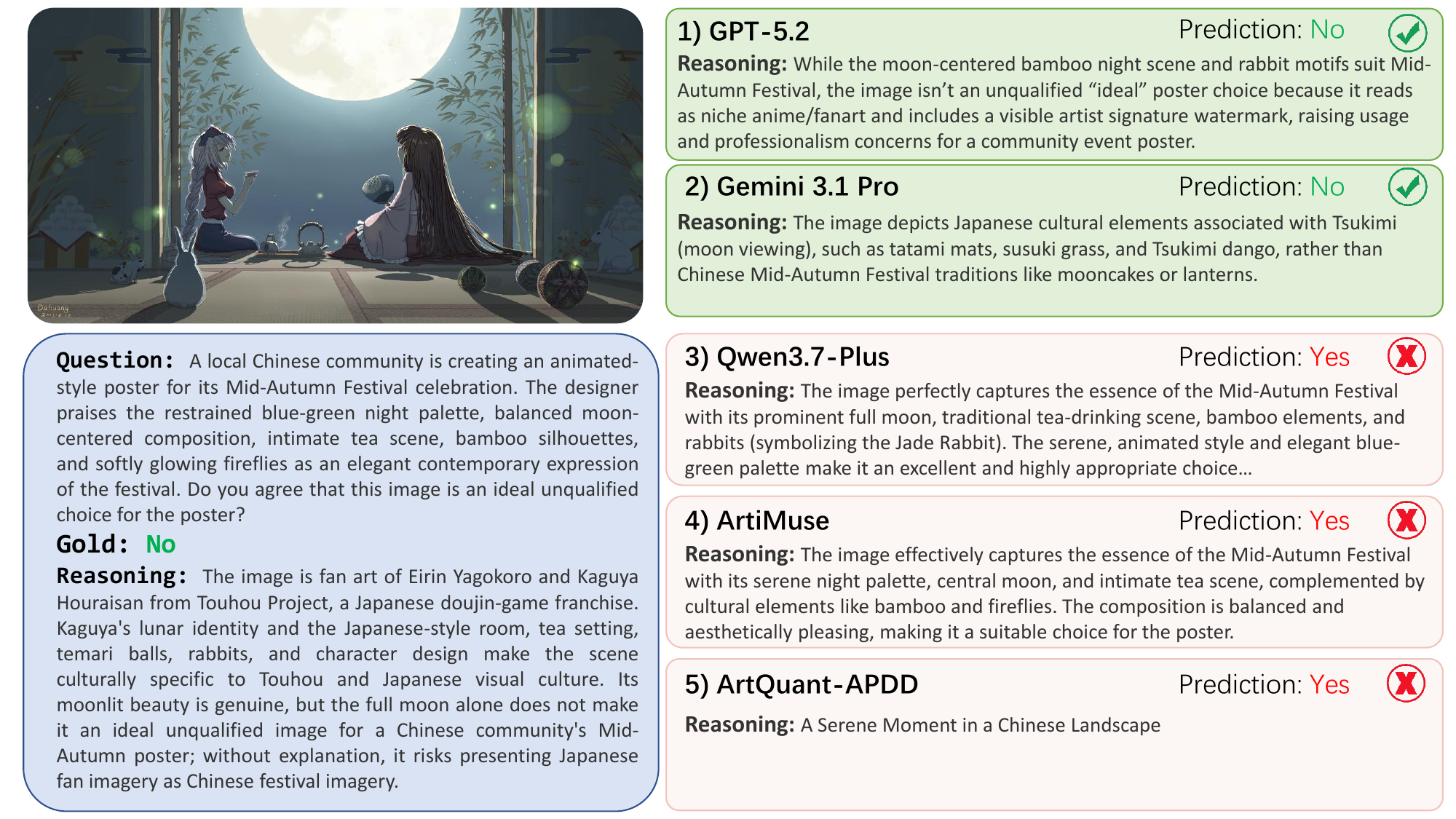}
\caption{Cross-model outputs for ContextCanvas case HB208 (gold: No).
GPT-5.2 and Gemini 3.1 Pro reject the image, whereas Qwen3.7-Plus,
ArtiMuse, and ArtQuant-APDD accept it. The displayed responses illustrate
whether models recognize the culturally specific visual identity rather than
relying only on surface festival cues.}
  \label{fig:hb208_outputs}
\end{figure}

\suppentry{Cultural-aesthetic alignment.}{Figure~\ref{fig:hb208_outputs}
compares five models on HB208. The image has an attractive moon-centered
composition, but its characters and visual language are tied to Japanese
Touhou fan culture rather than an unqualified Chinese Mid-Autumn identity.
Gemini 3.1 Pro and GPT-5.2 reject the image. Qwen3.7-Plus instead maps the full
moon, tea, bamboo, and rabbits directly to Chinese Mid-Autumn symbolism.
ArtiMuse repeats the positive formal framing and accepts it. ArtQuant-APDD also predicts Yes, accompanied only by the title-like phrase
``A Serene Moment in a Chinese Landscape,'' which misidentifies the cultural
setting. The disagreement is therefore not whether the image is
appealing, but whether that aesthetic reading is culturally valid for the
stated use.}

\begin{figure}[t]
  \centering
  \includegraphics[width=\textwidth]{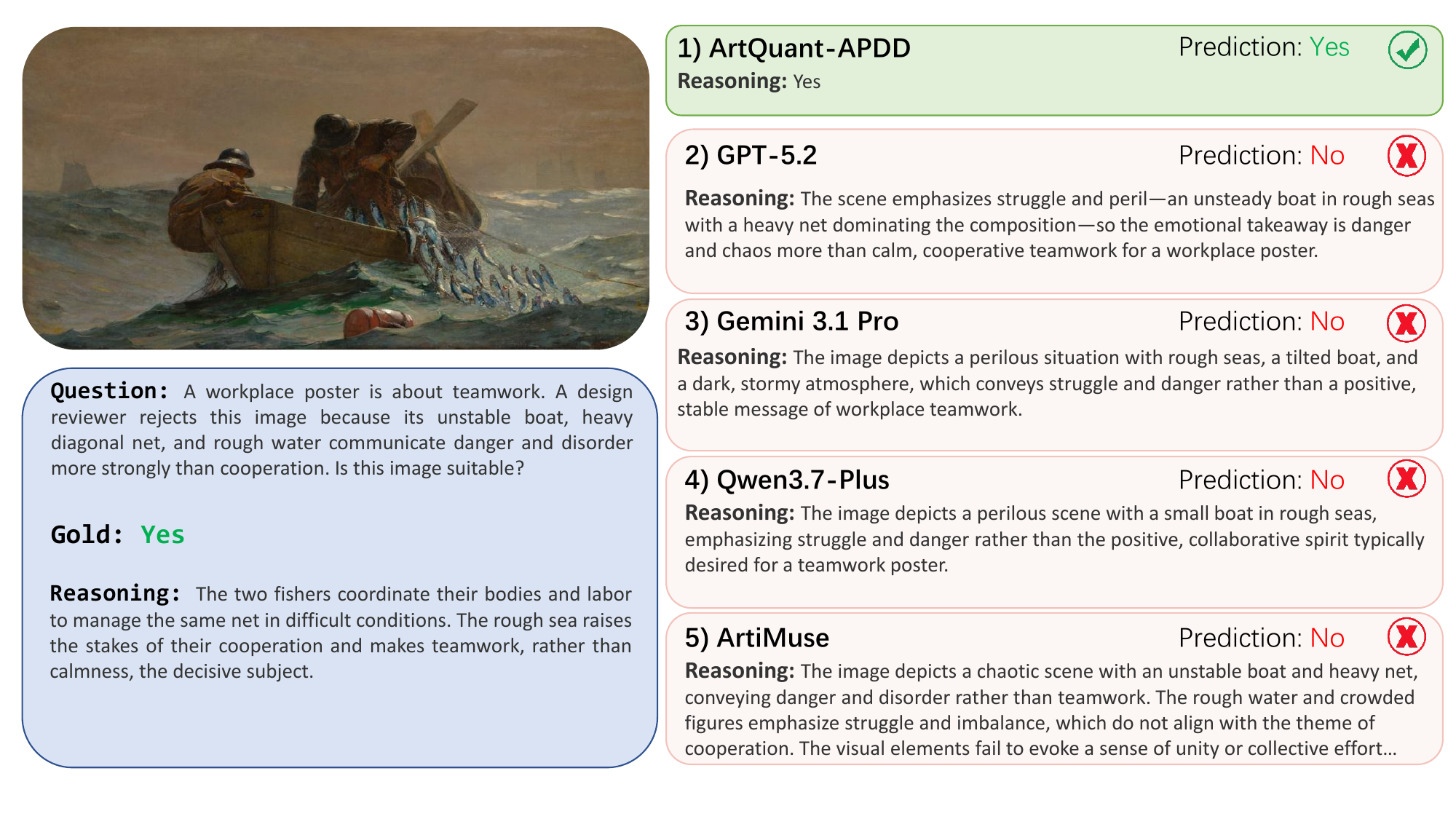}
  \caption{ArtQuant-APDD returns a bare Yes response without an explanatory rationale;}
  \label{fig:hb181_outputs}
\end{figure}

\suppentry{Aesthetic valence versus communicative function.}{Figure~\ref{fig:hb181_outputs}
compares five models on HB181, which tests whether models can distinguish
pleasant appearance from effective visual rhetoric.
Several models reject the painting because its rough water, dark tonal range,
tilted boat, and forceful diagonal net produce instability and tension rather
than a conventionally uplifting image. Yet these formal qualities intensify
the visible coordination of the two workers: their aligned bodies and shared
control of the net make cooperation legible precisely under pressure. The case
shows that contextual aesthetic suitability may depend on how composition,
tone, and visual tension serve a communicative purpose, rather than on positive
surface affect alone.}

\begin{figure}[t]
  \centering
  \includegraphics[width=\textwidth]{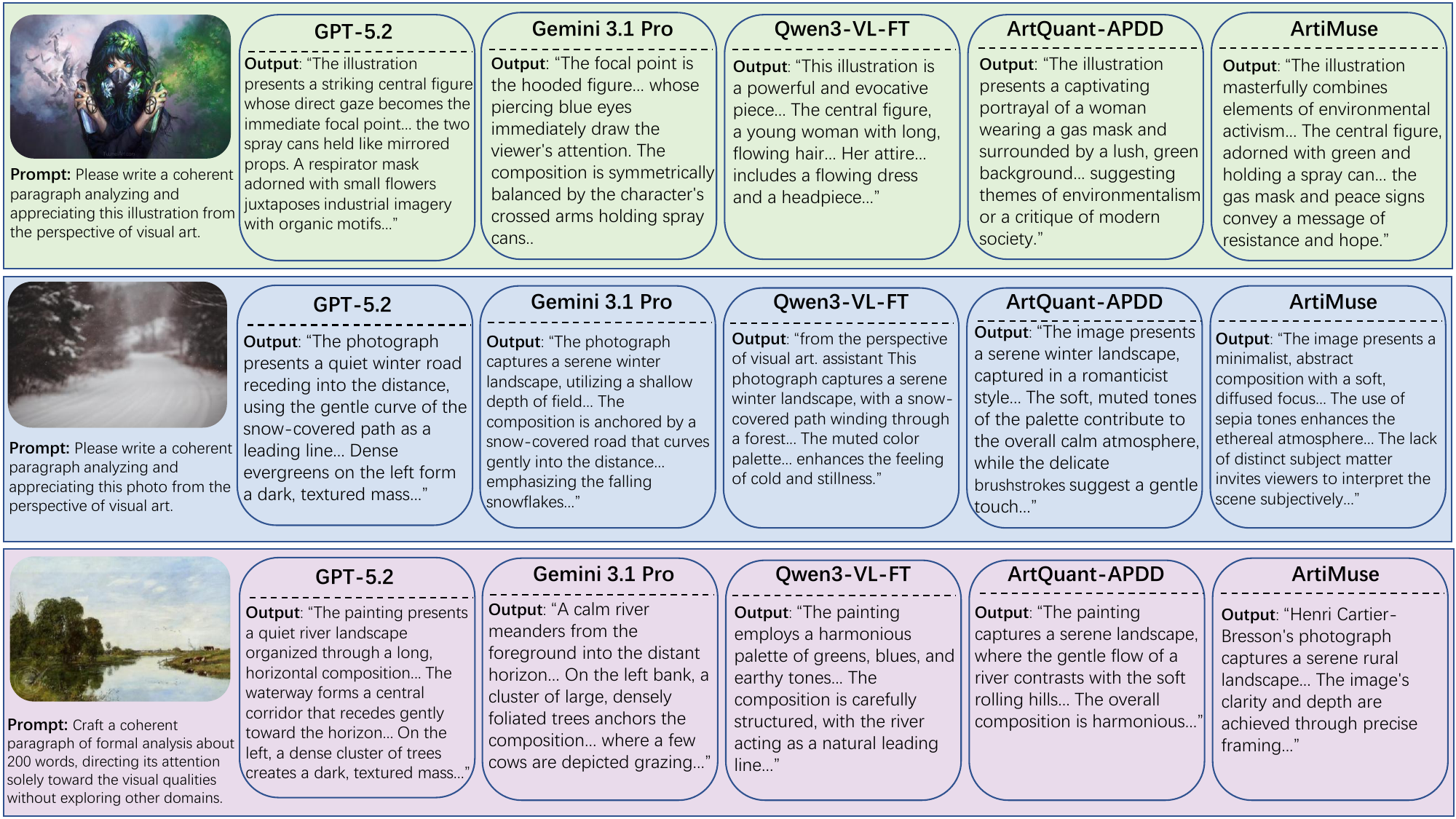}
  \caption{Three CritiqueCanvas cases compared across five models. From top to
  bottom: CC001 (illustration), CC003 (photograph), and CC002 (painting).
  The panel preserves each prompt and representative output excerpt, exposing
  differences in visual grounding, medium specificity, and analytical focus.}
  \label{fig:critique_crossmodel}
\end{figure}

\suppentry{Critique quality.}{Figure~\ref{fig:critique_crossmodel} supplies a
complementary generation-side comparison. Valid critiques of the same image may emphasize
different visual evidence or aesthetic criteria, so low overlap with one
reference does not necessarily imply low practical quality. The three selected
cases contrast grounded image-specific analysis with hallucinated subjects,
medium-inappropriate criteria, and fluent but generic interpretations.}

\subsubsection{Failure Modes and Representative Outputs}

The qualitative outputs instantiate the four failure levels introduced in the
main paper. \emph{Fluent but visually ungrounded} critiques invent objects or
properties absent from the image. \emph{Domain-blind quality priors} apply a
photographic defect vocabulary, such as blur or noise, to intentional painterly
texture. \emph{Surface aesthetics over communicative fit} treats attractive or
unattractive affect as sufficient evidence of suitability. \emph{Stylistic
plausibility over contextual fit} accepts an image because its palette,
composition, or period atmosphere appears coherent while overlooking the
identity, convention, or narrative that determines the use.

The ContextCanvas casebook further illustrates three recurrent subtypes without
claiming corpus-wide frequencies. In \emph{cultural-neighbor substitution}, a
nearby visual convention is mapped to the wrong occasion or identity (HB208,
HB014). In \emph{aesthetic-halo errors}, refinement or spectacle is treated as
proof of contextual appropriateness (HB037, HB186). In \emph{salient-composition
myopia}, a dominant formal cue suppresses a smaller but decisive action or
relation (HB106, HB181). These labels summarize reviewed examples; a frequency
claim would require an independent coding pass over all 301 cases.

\begin{figure}[t]
  \centering
  \includegraphics[width=\textwidth]{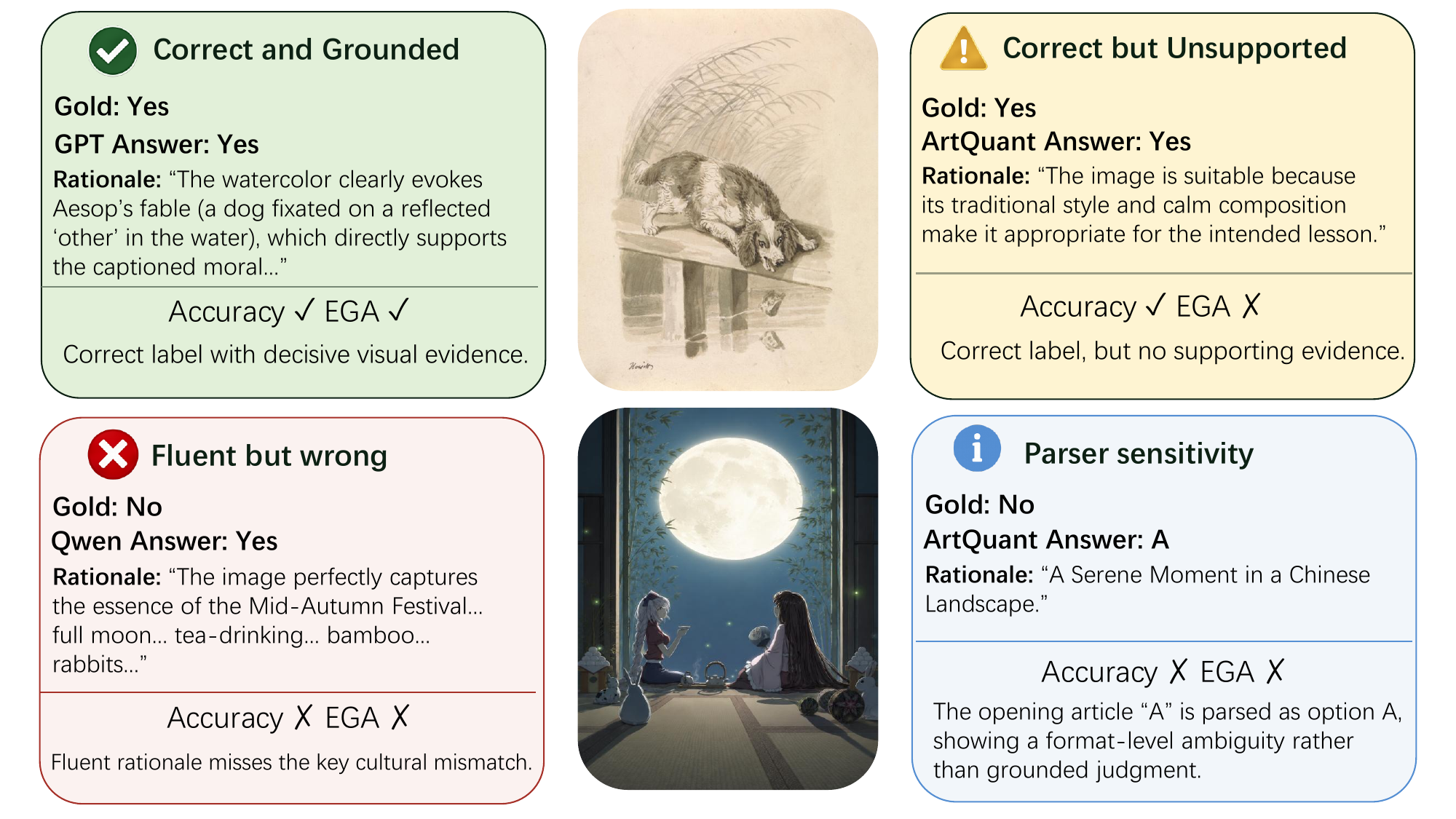}
  \caption{Response-level scoring diagnostics on E004 (top) and HB208
(bottom). E004 contrasts a grounded GPT-5.2 rationale with an unsupported
ArtQuant-APDD label. HB208 contrasts fluent but incorrect Qwen3.7-Plus
reasoning with an ArtQuant-APDD free-form response that illustrates
output-format and parser sensitivity.}
  \label{fig:output_types}
\end{figure}

Figure~\ref{fig:output_types} distinguishes a correct grounded response,
a correct but unsupported label, a fluent but incorrect rationale, and a
free-form response illustrating output-format sensitivity.

\subsection{Extended and Robustness Analyses}

\subsubsection{Extended Main Results}

The binary
error decomposition in Table~\ref{tab:classwise_context} explains why
Macro-F1 is lower than exact-match accuracy for many models: the gold set is
No-majority, while several models disproportionately predict Yes. In
particular, matching the marginal gold Yes rate is not sufficient---Grok's
37.80\% predicted Yes rate is close to 38.14\%, but its Macro-F1 is only
39.09\%. Contextual competence depends on instance-level alignment rather than
calibration of the marginal label count.

The main paper also promises response-level statistics beyond
reference-matching metrics. Table~\ref{tab:text_stats_supp} reports output
length, average dimension coverage, dimension density, adherence to
dimension-focused prompts, and the most frequent dimensions in otherwise
generic critiques. As defined in Section~\ref{sec:dataset_eval_materials},
these are descriptive measures of response form and explicit coverage rather
than direct measures of correctness or critique quality.

\begin{table}[t]
\centering
\small
\setlength{\tabcolsep}{3pt}
\caption{Generation style and aesthetic-dimension coverage on
CritiqueCanvas. Len/Ref is average output length relative to the reference;
DimHits is the average number of explicitly addressed dimensions;
aesthetic-term density is the number of matched lexical/phrase occurrences per
100 words. Prompt Align requires every dimension detected in a focused prompt
to be detected in the response.
Generic Top 3 excludes prompts that explicitly request a dimension.}
\label{tab:text_stats_supp}
\begin{tabular}{p{0.16\textwidth} c c c c c p{0.31\textwidth}}
\toprule
\textbf{Model} & \textbf{Length} & \textbf{Len/Ref} & \textbf{DimHits} &
\textbf{Density} & \textbf{Prompt Align} & \textbf{Generic Top 3 Dimensions} \\
\midrule
\multicolumn{7}{l}{\textit{Closed-source Models}} \\
Gemini 3.1 Pro  & 166.58 & 1.241 & \textbf{6.53} & 11.87 & 90.8\% & Composition, Emotion, Lighting \\
GPT-5.2         & 141.52 & 1.054 & 5.71 & 8.60  & 93.8\% & Composition, Emotion, Lighting \\
\midrule
\multicolumn{7}{l}{\textit{General MLLMs}} \\
GLM-4.6-Flash   & 189.37 & 1.410 & 6.50 & 9.44  & 94.2\% & Composition, Content, Emotion \\
InternVL3.5-8B  & 142.78 & 1.090 & 5.17 & 11.18 & 92.5\% & Content, Composition, Emotion \\
LLaVA-OneVision & 88.97  & 0.679 & 4.12 & 10.60 & 79.8\% & Content, Composition, Emotion \\
Qwen3-VL        & \textbf{199.07} & \textbf{1.520} & 5.54 & 9.29 & \textbf{95.2\%} & Composition, Content, Emotion \\
Qwen3-VL-FT     & 120.74 & 0.922 & 4.45 & 10.96 & 93.3\% & Composition, Content, Emotion \\
\midrule
\multicolumn{7}{l}{\textit{Aesthetic Models}} \\
ArtQuant   & 103.86 & 0.793 & 3.99 & 9.95  & 66.7\% & Content, Emotion, Composition \\
ArtiMuse   & 100.07 & 0.764 & 4.95 & 12.54 & 90.8\% & Composition, Content, Emotion \\
UniPercept & 102.12 & 0.780 & 5.15 & \textbf{13.26} & 92.3\% & Composition, Content, Emotion \\
AesExpert  & 35.68  & 0.272 & 2.34 & 10.34 & 34.4\% & Composition, Lighting, Color \\
Q-SiT      & 82.03  & 0.626 & 3.63 & 9.23  & 75.4\% & Composition, Content, Brushstrokes \\
\bottomrule
\end{tabular}
\end{table}

The statistics expose three complementary patterns. First, response length is
not a proxy for useful aesthetic analysis: Qwen3-VL produces the longest
critiques (199.07 words under the evaluation tokenizer), whereas UniPercept is substantially shorter
(102.12) but has the highest dimension density (13.26). Second, breadth and
controllability vary independently. Gemini obtains the broadest dimension
coverage (6.53 dimensions), while Qwen3-VL has the strongest focused-prompt alignment
(95.2\%); by contrast, ArtQuant and AesExpert follow such prompts much less
reliably (66.7\% and 34.4\%). Third, generic critiques concentrate heavily on
composition, content, and emotion across model families. Lighting, color, and
brushstrokes appear among the top dimensions for only a few models, while
technique, symbolism, and style never enter the generic top three. Thus,
models differ markedly in verbosity and response concentration, but their
open-ended critiques still rely on a relatively narrow set of salient
aesthetic concepts. These patterns complement---rather than replace---the
direct quality audit below, since greater length or dimension coverage does
not establish that the corresponding claims are relevant, correct, or
visually grounded.

\subsubsection{Critique Metric Validity}

Using the 100-case audit protocol described in
Section~\ref{sec:critique_eval_protocol},
Table~\ref{tab:critique_quality_extended} reproduces the aggregate direct
ratings reported in the main paper.

\begin{table}[H]
\centering
\caption{Aggregate human and MLLM ratings of critique quality on 100 randomly
sampled CritiqueCanvas cases.}
\label{tab:critique_quality_extended}
\small
\setlength{\tabcolsep}{7pt}
\begin{tabular}{lrr}
\toprule
\textbf{Model} & \textbf{Human} & \textbf{MLLM judge} \\
\midrule
GPT-5.2          & 3.9 & 5.0 \\
Gemini 3.1 Pro   & 4.1 & 4.5 \\
Qwen3-VL-FT      & 2.5 & 2.5 \\
ArtiMuse         & 2.6 & 2.6 \\
ArtQuant-APDD    & 2.3 & 2.2 \\
\bottomrule
\end{tabular}
\end{table}

The human ordering is consistent with the complementary judge at the group
level: GPT-5.2 and Gemini receive higher direct-quality ratings than the
adapted or aesthetic-specific models, although they do not dominate all
reference-similarity metrics. This supports a narrow conclusion. Lexical and
semantic reference alignment captures similarity to one expert answer but only
partially reflects grounding, image specificity, criterion choice, and
practical usefulness when multiple analyses are defensible. It does not show
that BLEU, SBERT, or CLIPScore is generally invalid. Figure~\ref{fig:critique_crossmodel} provides illustrative examples of this
mismatch: critiques with lower reference similarity can remain visually
grounded and analytically useful, whereas closer stylistic alignment does not
necessarily prevent generic analysis or unsupported visual claims.

\subsubsection{Counterfactual and Grounding Analyses}

\paragraph{Visual counterfactual sensitivity.}
Controlled visual counterfactuals can diagnose reliance on decisive image
evidence \citep{chen2020counterfactual}. We construct 36 matched No-to-Yes pairs by replacing a decisive incompatible
visual cue with a scenario-compatible alternative while keeping the use
scenario fixed.

Table~\ref{tab:counterfactual_extended} reports original and edited accuracy,
Pair Accuracy, directional updates, and the NCU defined in
Section~\ref{sec:context_eval_protocol}. Pair Accuracy requires both the
incompatible original and compatible edit to be classified correctly.

\begin{table}[H]
\centering
\caption{Counterfactual audit on 36 pairs. Full and
Orig. are accuracies on the complete benchmark and selected incompatible
originals; Edit is accuracy on compatible edits. Pair requires both labels to
be correct. C/R are correct and reverse directional updates. NCU confidence
intervals are paired bootstrap intervals.}
\label{tab:counterfactual_extended}
\scriptsize
\setlength{\tabcolsep}{3.5pt}
\begin{tabular}{lrrrrrrr}
\toprule
\textbf{Model} & \textbf{Full} & \textbf{Orig.} & \textbf{Edit} &
\textbf{Pair} & \textbf{C/R} & \textbf{NCU} & \textbf{95\% CI} \\
\midrule
GPT-5.2        & 71.43 & 72.22 & 58.33 & 36.11 & 13/2 & 30.56 & [11.11, 50.00] \\
Gemini 3.1 Pro & 85.71 & 86.11 & 38.89 & 27.78 & 10/1 & 25.00 & [8.33, 41.67] \\
Qwen3.7-Plus   & 51.83 & 52.78 & 75.00 & 30.56 & 11/1 & 27.78 & [11.11, 44.44] \\
\midrule
ArtQuant-APDD  & 29.57 & 30.56 & 72.22 &  8.33 &  3/2 &  2.78 & [-8.33, 13.89] \\
ArtiMuse       & 19.60 & 19.44 & 97.22 & 16.67 &  6/0 & 16.67 & [5.56, 30.56] \\
\bottomrule
\end{tabular}
\end{table}

The three general-purpose models occupy a narrow 25.00--30.56 NCU range despite
substantially different original accuracy. ArtQuant changes its decision on
only five pairs, with three correct and two reverse updates, producing an NCU
of 2.78 whose interval includes zero. ArtiMuse makes six correct and no reverse
updates, but predicts Yes for 64 of the 72 original/edit images. Its 97.22\%
edited accuracy therefore largely reflects an acceptance tendency; Pair
Accuracy and NCU provide the more discriminative diagnosis. The mean NCU is
27.78 for the three general-purpose models and 9.72 for the two specialists.

The rationale audit additionally reveals source-scene anchoring: some responses
repeat a canonical narrative even after its supporting visual cue is removed.

\begin{figure}[t]
  \centering
  \includegraphics[width=\textwidth]{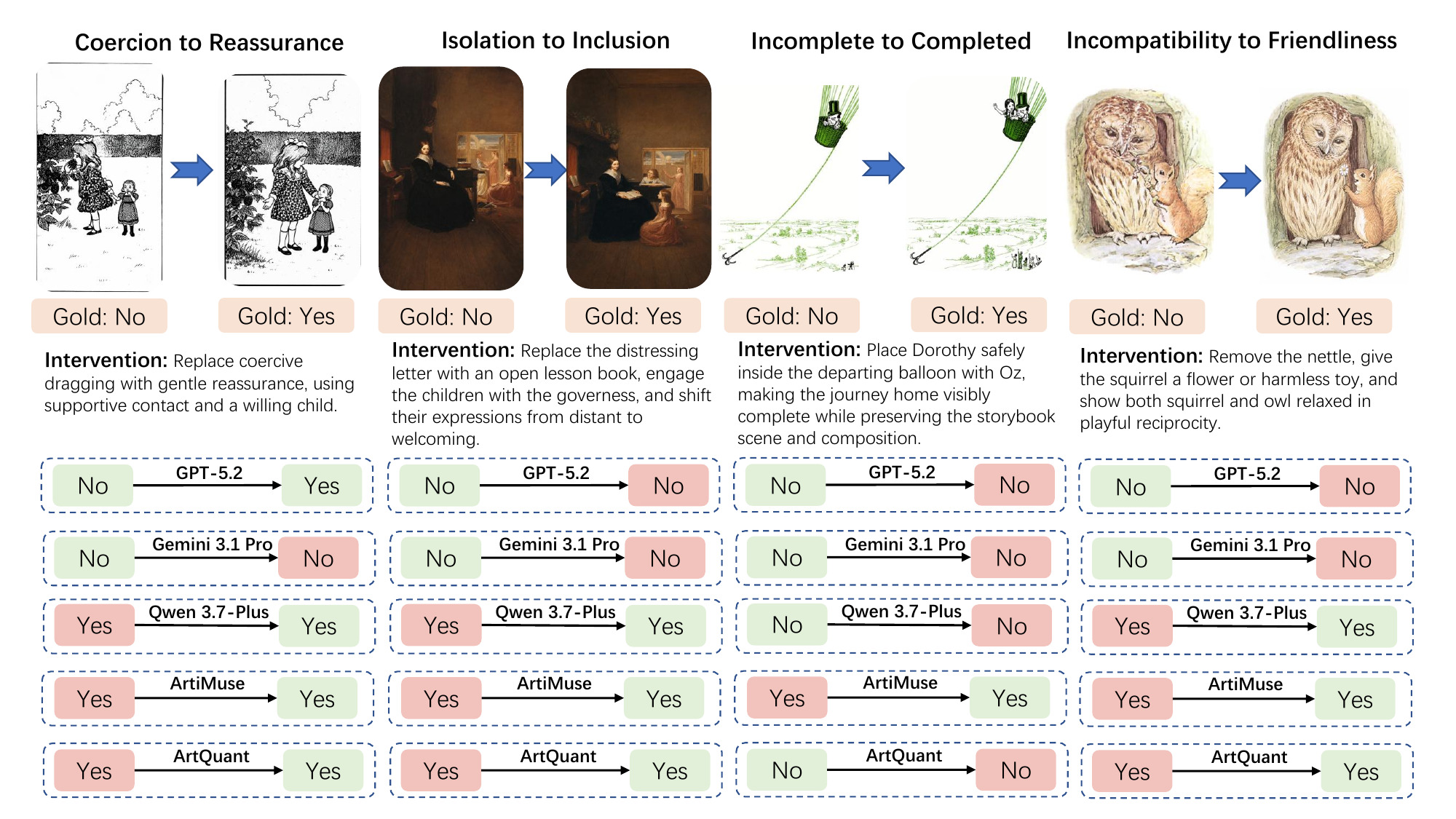}
  \caption{Four No-to-Yes counterfactual interventions. From left to right:
  CF20/HB163 (coercion to reassurance), CF18/HB084 (isolation to inclusion),
  CF28/E108 (incomplete to completed journey), and CF35/HB075 (incompatibility
  to friendliness). Each column shows the original and edited image, the
  targeted intervention, and each model's before--after prediction; colored
  boxes indicate agreement or disagreement with the corresponding gold label.}
  \label{fig:counterfactual_examples}
\end{figure}

Figure~\ref{fig:counterfactual_examples} presents representative interventions
with each model's before--after prediction.

\paragraph{Evidence-grounded accuracy.}
Using the EGA and Unsupported definitions in
Section~\ref{sec:context_eval_protocol},
Table~\ref{tab:ega_extended} expands the 100-case audit.

\begin{table}[H]
\centering
\caption{Decisive-cue grounding on 100 ContextCanvas cases.}
\label{tab:ega_extended}
\small
\setlength{\tabcolsep}{7pt}
\begin{tabular}{lrrr}
\toprule
\textbf{Model} & \textbf{Accuracy} & \textbf{EGA} & \textbf{Unsupported (\%)} \\
\midrule
Gemini 3.1 Pro & 83 & 79 & 4.82 \\
GPT-5.2        & 65 & 55 & 15.38 \\
Qwen3.7-Plus   & 42 & 31 & 26.19 \\
ArtiMuse       & 21 &  3 & 85.71 \\
ArtQuant-APDD & 27 & 0 & 100.00 \\
\bottomrule
\end{tabular}

\end{table}

ArtiMuse is the cleaner specialist diagnosis because it always generates a
substantive rationale: its fall from 21 correct labels to 3 grounded correct
answers cannot be attributed to missing explanation text. Its prose is often
fluent and aesthetically framed, but it seldom identifies and correctly binds
the culture- or context-bearing cue. The audit therefore addresses the concern
that binary accuracy alone may reward lucky labels or unsupported agreement.

\paragraph{Perception--reasoning decomposition.}
Following the view of captions as task-facing visual surrogates
\citep{yang2026captionqa}, Table~\ref{tab:modality_decomposition} evaluates a
100-case diagnostic subset distinct from the 100 cases used for the
evidence-grounding audit.
Four conditions are compared: question and options
only (\emph{Text}); an identity- and verdict-free description of visible
content (\emph{Neutral}); the same description with the decisive visible cue
made explicit (\emph{Evidence}); and the original image (\emph{Image}).

\begin{table}[H]
\centering
\caption{Accuracy under different visual-information conditions on 100
ContextCanvas cases.}
\label{tab:modality_decomposition}
\small
\setlength{\tabcolsep}{6pt}
\begin{tabular}{lrrrr}
\toprule
\textbf{Model} & \textbf{Text} & \textbf{Neutral} &
\textbf{Evidence} & \textbf{Image} \\
\midrule
GPT-5.2        & 45 & 77 & 84 & 71 \\
Gemini 3.1 Pro & 58 & 86 & 90 & 86 \\
Qwen3.7-Plus   & 44 & 78 & 88 & 52 \\
\midrule
ArtQuant-APDD  & 16 & 20 & 19 & 27 \\
ArtiMuse       & 16 & 32 & 47 & 20 \\
\bottomrule
\end{tabular}
\end{table}

Text-only performance does not exceed the 61\% majority baseline. Supplying a
neutral visual description improves the three general-purpose models by
28--34 points; for each model, the paired exact two-sided McNemar test gives
$p<10^{-7}$. This confirms that the judgments
depend materially on visual content. The modality gap is model-specific:
Gemini matches its image accuracy from a neutral description, whereas Qwen
rises from 52\% with images to 78\% with neutral descriptions and 88\% when
the decisive cue is explicit. ArtiMuse is partly rescued by explicit evidence
but remains far below the general-purpose models; ArtQuant remains weak across
all conditions. Together with the counterfactual and EGA results, these
conditions provide a diagnostic of whether errors are more consistent with
limited visual-evidence extraction or with failure to use explicitly supplied
contextual evidence; they are not a definitive causal decomposition.

\subsubsection{Base-to-Specialist Transfer}

Table~\ref{tab:lineage_extended} compares four aesthetic specialists with the
base checkpoint identified by their implementation and model documentation. Because every
pair is evaluated on the same 301 cases, we report the two discordant counts,
a paired bootstrap interval for the accuracy change, an exact two-sided
McNemar test, and the Bonferroni-adjusted value across four lineages.

\begin{table}[H]
\centering
\caption{Paired base-to-specialist ContextCanvas transfer. $b$ counts cases
correct only for the base and $c$ cases correct only for the specialist. Raw
McNemar $p$-values and Bonferroni-adjusted values are both shown.}
\label{tab:lineage_extended}
\scriptsize
\setlength{\tabcolsep}{3.2pt}
\begin{tabular}{lrrrrrr}
\toprule
\textbf{Base $\rightarrow$ specialist} & \textbf{Base} & \textbf{Spec.} &
\textbf{$\Delta$} & \textbf{$b/c$} & \textbf{95\% CI} &
\textbf{$p/p_{\mathrm{adj}}$} \\
\midrule
mPLUG-Owl2 $\rightarrow$ ArtQuant & 20.27 & 29.57 & +9.30 & 29/57 & [3.32, 15.28] & .0034/.0134 \\
InternVL3-8B $\rightarrow$ ArtiMuse & 23.26 & 19.60 & -3.65 & 27/16 & [-7.97, 0.33] & .1263/.5052 \\
LLaVA-v1.5 $\rightarrow$ AesExpert & 12.62 & 18.27 & +5.65 & 10/27 & [1.66, 9.63] & .0076/.0305 \\
LLaVA-OV $\rightarrow$ Q-SiT & 19.93 & 19.60 & -0.33 & 12/11 & [-3.32, 2.66] & 1.000/1.000 \\
\bottomrule
\end{tabular}
\end{table}

Specialization has a heterogeneous effect. ArtQuant and AesExpert significantly
improve over their corresponding bases after correction, while ArtiMuse and
Q-SiT show no measurable ContextCanvas gain. Thus, specialization does not
produce consistent gains in contextual judgment.

\bibliographystyle{aaai2027}
\bibliography{ref}